\pdfoutput=1
\documentclass[11pt]{article}

\usepackage[preprint]{acl}

\usepackage{times}
\usepackage{latexsym}

\usepackage[T1]{fontenc}
\usepackage[utf8]{inputenc}

\usepackage{microtype}
\usepackage{booktabs}
\usepackage{multicol}
\usepackage{tikz}
\usepackage{subcaption}
\usepackage{inconsolata}

\usepackage{graphicx}
\usepackage{amsmath, amsfonts}
\usepackage{cleveref}

\usepackage{tabularx}
\usepackage{csquotes}
\usetikzlibrary{positioning, calc}
\usepackage{balance}

\title{The Language of Trauma: Modeling Traumatic Event Descriptions Across Domains with Explainable AI}

\author{Miriam Schirmer\textsuperscript{1}, Tobias Leemann\textsuperscript{1}, Gjergji Kasneci\textsuperscript{1}, Jürgen Pfeffer\textsuperscript{1}, David Jurgens\textsuperscript{2} \\
\textsuperscript{1}Technical University of Munich \\
\textsuperscript{2}University of Michigan \\
}

\begin{document}
\maketitle
\begin{abstract}
Psychological trauma can manifest following various distressing events and is captured in diverse online contexts. However, studies traditionally focus on a single aspect of trauma, often neglecting the transferability of findings across different scenarios. We address this gap by training language models with progressing complexity on trauma-related datasets, including genocide-related court data, a Reddit dataset on post-traumatic stress disorder (PTSD), counseling conversations, and Incel forum posts.
Our results show that the fine-tuned RoBERTa model excels in predicting traumatic events across domains, slightly outperforming large language models like GPT-4. Additionally, SLALOM-feature scores and conceptual explanations effectively differentiate and cluster trauma-related language, highlighting different trauma aspects and identifying sexual abuse and experiences related to death as a common traumatic event across all datasets.
This transferability is crucial as it allows for the development of tools to enhance trauma detection and intervention in diverse populations and settings.

\end{abstract}

\section{Introduction}

Post-Traumatic Stress Disorder (PTSD) is a significant mental health condition that can develop after experiencing a traumatic event. For an event to potentially lead to PTSD, it must involve actual or threatened death, serious injury, or a threat to one's physical integrity, causing intense fear, helplessness, or horror \citep{friedman2007handbook,gold2017apa}. Although about 70\% of Americans will encounter such traumatic events in their lifetime, only about 5-7\% develop PTSD, highlighting that PTSD is relatively rare despite high trauma exposure. However, this figure could be higher, as many cases may go undiagnosed \citep{bonn2022long, atwoli2015epidemiology}.

This discrepancy suggests that various factors, including psychological resilience, the nature of the trauma, and access to mental health support, influence the development of PTSD. Definitions of trauma and responses to it can vary widely across cultures and social contexts, affecting the prevalence and expression of PTSD.

\begin{figure}[t!]
\centering
\resizebox{\columnwidth}{!}{
\begin{tikzpicture}
    \definecolor{goldenYellow}{RGB}{255,239,153}
    \definecolor{orange}{RGB}{255,204,153}
    \definecolor{lightCoral}{RGB}{255,182,193}
    % Frame around the whole figure
    \node[draw, rounded corners, thick, inner sep=0.4cm] (frame) at (0,0) {

    \begin{tikzpicture}[every node/.style={inner sep=0,outer sep=0}]
        % Data Sources
        \node[draw, fill=orange, minimum width=4.4cm, minimum height=0.9cm, align=center] (rect1) at (0,0) {\textbf{Genocide} Tribunals};
        \node[draw, fill=orange, minimum width=4.4cm, minimum height=0.9cm, align=center, right=of rect1] (rect2) {\textbf{PTSD} Reddit};
        \node[draw, fill=orange, minimum width=4.4cm, minimum height=0.9cm, align=center, right=of rect2] (rect3) {\textbf{Counseling} Conversations};
        \node[draw, fill=orange, minimum width=4.4cm, minimum height=0.9cm, align=center, right=of rect3] (rect4) {\textbf{Incel} Posts};

        \node[align=center, above=0.9cm of $(rect2)!0.5!(rect3)$, inner sep=8pt] (caption) {\Large \textbf{Data Sources}};

        % Binary Classification Task
        \node[below=2.2cm of $(rect2)!0.5!(rect3)$, minimum width=12cm, align=center, inner sep=10pt] (middle) {\Large \textbf{Binary Classification Task with Multiple Models}: \\ Does the Text Contain a Traumatic Event?};

        % Rectangles for Trauma Event and No Trauma Event
        \node[draw, fill=goldenYellow, minimum width=4cm, minimum height=0.9cm, align=center, below=1.2cm of middle, xshift=-4cm] (trauma) {Trauma Event};
        \node[draw, fill=goldenYellow, minimum width=4cm, minimum height=0.9cm, align=center, below=1.2cm of middle, xshift=4cm] (no_trauma) {No Trauma Event};

        \draw[->] ($(rect2.south)!0.5!(rect3.south)$) -- ++(0,-0.05) -- ++(0,-0.15) -- (middle.north);

        % Arrows to new rectangles
        \draw[->] (middle.south) -- ++(0,-0.4) -- (trauma.north);
        \draw[->] (middle.south) -- ++(0,-0.4) -- (no_trauma.north);

        % XAI Methods
        \node[below=2.9cm of middle, align=center, inner sep=20pt] (bottom) {\Large \textbf{XAI Methods to find Common Characteristics} \\ Conceptual Explanations and Feature Importance Scores}; % Increased space here
        \draw[->] (trauma.south) -- ++(0,-0.4) -- (bottom.north);
        \draw[->] (no_trauma.south) -- ++(0,-0.4) -- (bottom.north);

        % Common Concepts and Features
        \node[draw, fill=lightCoral, minimum width=2cm, minimum height=1cm, align=center, below=0.4cm of bottom, xshift=-3.6cm] (overlap1) {concept};
        \node[draw, fill=lightCoral, minimum width=2cm, minimum height=1cm, align=center, right=-0.5cm of overlap1] (overlap2) {concept};
        \node[draw, fill=lightCoral, minimum width=2cm, minimum height=1cm, align=center, right=-0.5cm of overlap2] (overlap3) {concept};
        \node[below=0.4cm of overlap2, align=center] (label1) {Overlapping Concepts};
        
        \node[draw, fill=lightCoral, circle, minimum size=1.2cm, align=center, below=0.4cm of bottom, xshift=1.4cm] (circle1) {feature};
        \node[draw, fill=lightCoral, circle, minimum size=1.2cm, align=center, right=-0.2cm of circle1] (circle2) {feature};
        \node[draw, fill=lightCoral, circle, minimum size=1.2cm, align=center, right=-0.2cm of circle2] (circle3) {feature};
        \node[below=0.4cm of circle2, align=center] (label2) {Overlapping Features};

    \end{tikzpicture}
    };
\end{tikzpicture}
}
\caption{We (1) create a cross-domain trauma dataset, (2) classify traumatic events with models of different complexity, and (3) use XAI methods to identify overlapping characteristics of traumatic events.}
\end{figure}
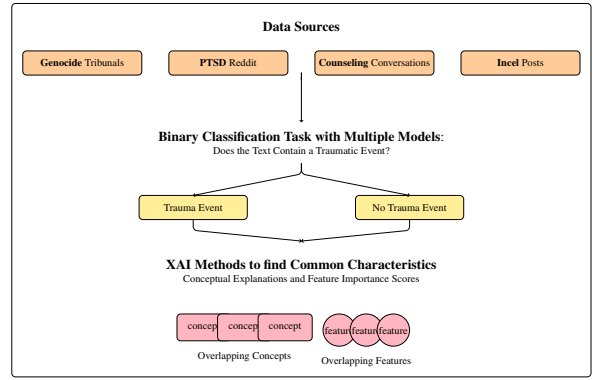

To investigate the interplay of these factors, we are proposing a Natural Language Processing (NLP) approach to identify traumatic events across different domains. Understanding the cross-cutting mechanisms of trauma is crucial for developing comprehensive support systems and interventions that are adaptable to various contexts.
We are following up on these research questions:

\textbf{RQ1:} Given the diverse forms of trauma, what are the most effective methods for modeling and predicting its manifestations?

\textbf{RQ2:} How transferable is the detection of multifaceted traumatic events across domains?

\textbf{RQ3:} What are the cross-cutting mechanisms related to trauma that can be identified across different types and contexts of traumatic events?

\vspace{0.2cm}

Our work advances trauma detection by applying NLP and XAI methods to offer detailed insights not yet explored in the literature. We contribute by: (1) identifying key trauma concepts from psychological literature and replicating them using NLP methods, (2) modeling traumatic event detection with various language models and creating a dataset that includes genocide court transcripts, PTSD-related Reddit posts, counseling conversations, and \enquote{Involuntary Celibates} Incel forum posts, (3) developing a three-stage XAI framework that approximates Shapley values, assesses feature importance, and identifies task-relevant concepts, providing a comprehensive understanding of trauma at both the instance and dataset levels, and (4) automating trauma detection to enhance online psychological support by displaying hotline information and resources in forums where trauma is frequently discussed.
%\footnote{We make all data and code available at \url{https://github.com/MiriamSchirmer/trauma-language}}

\section{Traumatic Events \& Language}

\subsection{Definition \& Scope}
Psychological trauma, as defined by the American Psychological Association (APA), encompasses experiences of "exposure to actual or threatened death, serious injury, or sexual violence," whether directly encountered or witnessed. This includes instances where individuals "learn that the traumatic event(s) occurred to a close family member or close friend" \citep{APA2013}. 

While psychological trauma and PTSD are frequently discussed in the context of childhood abuse and the military, trauma can manifest in a variety of situations \citep{van2003psychological, yehuda1998psychological}. It can arise in interpersonal violence like domestic abuse and sexual assault; and accidents or natural disasters. Trauma can also result from medical issues, bereavement and loss, emotional and psychological abuse, and its manifestation can vary depending on cultural beliefs and values \citep{smelser2004psychological}. 
%Secondary trauma affects first responders, healthcare workers, and therapists exposed to others' traumatic experiences \citep{elwood2011secondary,newell2010professional}. 
%Personal history, support systems, the nature of the trauma, and coping mechanisms influence individual responses to trauma.

\subsection{Trauma Contexts \& Categorization}

Within the psychological literature, key events have been identified that are typical for specific trauma contexts. 
In armed conflict and mass atrocities, exposure to severe violence and death is prevalent. This often includes the death of close family members, forced displacement, and sexual abuse \citep{powell2003posttraumatic}. For instance, \citet{dyregrov2000trauma} found that most child survivors of the Rwandan genocide had witnessed severe injuries and deaths, with more than half witnessing massacres.

In domestic trauma, the most common forms are physical abuse (e.g., intimate partner violence), emotional abuse, and neglect \citep{mccloskey2000posttraumatic}. Emotional abuse is particularly hard to detect due to its subtle nature, including consistent belittling, criticizing, or bullying \citep{dye2020emotional,idsoe2021bullying}.
%These forms of trauma can manifest in various ways, such as children experiencing chronic fear, anxiety, depression, and difficulties in forming healthy relationships. 
%Witnessing domestic violence is another significant trauma, especially for children, leading to long-term psychological issues.
Sexual violence, whether in war or domestic contexts, is an especially devastating form of trauma \citep{kiser1991physical}. This includes childhood sexual abuse, rape, and exploitation.
%Each type of sexual abuse can result in severe psychological impacts, including PTSD, depression, anxiety, and issues with self-esteem and trust. 
%Survivors of childhood sexual abuse, in particular, often struggle with long-term emotional and psychological scars that affect their development and relationships.

The range of traumatic events makes conceptualizations of trauma complex. Researchers have categorized trauma in line with diagnostic manuals like the Diagnostic and Statistical Manual of Mental Disorders (DSM) into types such as assaultive violence (e.g., military combat, rape, threats with weapons), other injuries or shocking events (e.g., serious car accidents and life-threatening illnesses) \citep{breslau2004estimating}.
%The nature of the traumatic event, whether witnessing mass killings or experiencing sexual abuse by a family member, significantly affects how trauma manifests and how survivors articulate their experiences. 
Identifying these events is crucial, as most subsequent issues are linked to the initial trauma due to the development of trauma-specific fears in PTSD \citep{terr2003childhood}.

\subsection{NLP for Trauma Detection}

Given the variety and subjective nature of traumatic experiences, detecting them in text is complex. Despite these challenges, recent research has shown that NLP methods can improve the detection of psychological disorders and aid in treatment adaptation \cite{ahmed2022machine, de2014mental, leglaz2021machine, malgaroli2023natural, zhang2022natural}. 

\noindent \textbf{NLP and Mental Health.} Major areas in this field include promoting better health and early disorder identification for intervention \citep{calvo2017natural,swaminathan2023natural}. For example, \citet{levis2021natural} associated linguistic markers from psychotherapist notes with treatment duration. Analyzing mental health chat conversations, \citet{hornstein2024predicting} found that words indicating younger age and female gender were associated with a higher chance of re-contacting.
%More generally, \citet{althoff2016large} developed a framework for text-message-based counseling to correlate various linguistic aspects with conversation outcomes. 

Recently, the use of Large Language Models (LLMs) has led to the development of specific models for mental health applications \citep{xu2024mental, yang2024mentallama}. While LLMs effectively detect mental health issues and provide eHealth services, their clinical use poses risks, such as the lack of expert-annotated multilingual datasets, interpretability challenges, and issues regarding data privacy and over-reliance \citep{guo2024large}.

Specifically for social media data, there has been research on using sentiment analysis and semantic structures to detect anxiety \citep{low2020natural} or depression \citep{tejaswini2024depression} on Reddit posts. In suicide prevention on social media, \citet{sawhney2020time} developed a superior model for suicidal risk screening that identifies emotional and temporal cues, outperforming competitive methods (c.f., \citet{ji2022towards} on suicidal risk detection).

\noindent  \textbf{Trauma Detection.} In trauma research, progress is being made in analyzing patient narratives \citep{he2017automated} and identifying cases of post-traumatic stress disorder (PTSD) through speech \citep{marmar2019speech}. \citet{miranda2024enhancing} developed an NLP workflow using a pre-trained transformer-based model to analyze clinical notes of PTSD patients, revealing consistent reductions in trauma criteria post-psychotherapy. Disruptions in lexical characteristics and emotional valence have been found to contribute to identifying PTSD \citep{quillivic2024interdisciplinary}.
%Specifically within genocide research, researchers have developed a model to identify potentially traumatic content within witness statements from international criminal courts \citep{schirmer2023uncovering}.
Using Twitter data, \citet{alam2020laxary} investigated whether posts can complete clinical PTSD assessments, achieving promising accuracy in PTSD classification and intensity estimation validated with veteran Twitter users (cf. \citet{coppersmith2014measuring, reece2017forecasting}).

\subsection{Trauma Event Detection in this Study}

%The effectiveness of detecting mental health phenomena depends on the conceptualization of the phenomenon and its identification through text analysis. Keywords and sentiments associated with certain semantic structures, such as the use of pronouns, play a crucial role \citep{low2020natural}. 

Previous work has identified language markers of PTSD, such as overuse of first-person singular pronouns, increased use of words related to depression, anxiety, and death, and more negative emotions. However, these markers are not specific to trauma and can also be associated with other psychological disorders, complicating accurate identification. Additionally, the transferability of detection methods is often lacking \citep{coppersmith2014measuring, quillivic2024interdisciplinary}.

Trauma detection in NLP is distinct in that it involves identifying a specific traumatic event that precedes a PTSD diagnosis, unlike the detection of depression or anxiety, which do not require a concrete event in their definitions. This study focuses on detecting such events in online resources, avoiding symptom or diagnosis analysis. Drawing conclusions about mental health from public text data alone is impossible without additional psychological information. We aim to identify instances meeting the APA's definition of trauma, minimizing subjectivity by closely following their criteria.

\begin{table*}[t]\small
    \centering
    \begin{tabularx}{\textwidth}{@{}>{\raggedright\arraybackslash}p{0.195\textwidth} p{0.43\textwidth} p{0.17\textwidth} p{0.2\textwidth}@{}}
        \toprule
        \textbf{Dataset} & \textbf{Description} & \textbf{Size \& Balance} & \textbf{AA} \\
        \midrule
        Genocide Transcript Corpus (GTC) & Witness statements from 90 different cases across three different genocide tribunals.  & 
        \begin{tabular}[t]{@{}l@{}}
            $15{,}845$ samples \\
            (trauma: 13.54\%) 
        \end{tabular} & n/a \\ % Added an extra cell for Annotator Agreement
        \midrule
        PTSD Subreddit (PTSD) & Post-Traumatic Stress Disorder (PTSD) subset of the Reddit Mental Health Dataset. & 
        \begin{tabular}[t]{@{}l@{}}
            $1{,}200$ samples \\
            (trauma: $47.19\%$) 
        \end{tabular} & 
        \begin{tabular}[t]{@{}l@{}}
            (1) $\alpha = .63$ \\
            (2) $F1 = .77$
            \end{tabular} \\
        \midrule
        Counseling Dataset & Queries submitted by users seeking advice, with answers provided by professionals. & 
        \begin{tabular}[t]{@{}l@{}}
            $1{,}200$ samples \\
            (trauma: $8.16\%$) 
        \end{tabular} & 
        \begin{tabular}[t]{@{}l@{}}
            (1) $\alpha   = .69$ \\
            (2) $F1 = .95$
            \end{tabular} \\
        \midrule
        Incel Dataset & Posts from the Incel online forum \emph{incels.is}. & 
        \begin{tabular}[t]{@{}l@{}}
            $300$ samples \\
            (trauma: $2.67$\%) 
        \end{tabular} & 
        \begin{tabular}[t]{@{}l@{}}
            (1) $\alpha  = .43$ \\
            (2) $F1 = .78$
            \end{tabular} \\
        \bottomrule
    \end{tabularx}
    \caption{Dataset Overview. Note: Annotator agreement (AA) was calculated (1) among crowd workers (Krippendorff's $\alpha$) and (2) for the crowd worker majority vote vs. the expert vote (Binary F1).}
    \label{tab:dataset_overview}
\end{table*}

%\subsection{Summary \& Contribution}
%Given the lack of a clear operationalization of traumatic language that is detectable in text alone, this study addresses three primary research questions: identifying the most effective methods for modeling and predicting the diverse manifestations of trauma, examining the transferability of detecting multifaceted traumatic events across different domains, and exploring the cross-cutting mechanisms related to trauma that can be identified across various types and contexts of traumatic events.

%To gain a comprehensive understanding of the text, we employ three distinct XAI methods: first, we use Shapley approximations on an instance level to understand the contribution of each feature to individual predictions; next, we extract common features that contribute to trauma across all datasets; and finally, we examine conceptual explanations for each trauma domain.

\section{Data \& Labeling}

\subsection{Data Sources}

 Our final dataset is built from four datasets, each offering unique perspectives on traumatic experiences (\Cref{tab:dataset_overview}) to identify common characteristics of trauma that extend beyond specific events, such as those related to war: The Genocide Court Transcripts (GTC;  \citealp{schirmer2023uncovering}) dataset comprises text from genocide tribunals, providing insights into severe human rights violations and the profound trauma experienced by victims and witnesses. This encompasses 90 cases across the International Criminal Tribunal for Rwanda, the International Criminal Tribunal for the former Yugoslavia, and the Extraordinary Chambers in the Courts of Cambodia. The Reddit PTSD Dataset includes posts from the PTSD subreddit of the Reddit Mental Health Dataset \cite{low2020natural}, where individuals discuss their experiences with post-traumatic stress disorder, sharing personal stories and support. The Mental Health Counseling Conversations Dataset \citep{amod_2024} features questions and answers sourced from online counseling and therapy platforms. The questions cover a wide range of mental health topics, and qualified psychologists provide the answers. 

The Incel Posts Dataset \citep{matter2024investigating} contains posts from Incel community forums and reflects extreme misogynistic viewpoints. This dataset serves as a control in our study: Though not explicitly trauma-related, it includes posts on depression, bullying, and violence directed towards women. The violent and aggressive language in this dataset helps quantify our models’ ability to distinguish explicit trauma from related emotional distress.

\subsection{The Trauma Event Dataset TRACE}
We present the final trauma event dataset TRACE (\underline{T}rauma Event \underline{R}ecognition \underline{A}cross \underline{C}ontextual \underline{E}nvironments).
To that end, all source datasets were pre-processed to ensure comparability for the detection task, including the removal of URLs and standardization of formatting. Due to their varied origins, the samples from each dataset differ in size, with instances ranging from single-word sentences to more elaborate descriptions of events and personal thoughts across all datasets.
For compatibility with the BERT-architecture, we split instances exceeding the 512-token limit into smaller segments. Our approach treats each segment as independent, with trauma classification based solely on its content. While some segments from the same text may appear in both training and test sets, we consider label leakage minimal, since the model must rely on the segment's content for accurate prediction. 7-20\% (depending on the dataset) of segments were split overall.

Our study aims to demonstrate cross-domain transferability on realistic data, making it crucial to use datasets with their expected class distribution, even if they differ in context and trauma event rates. We matched the size of all datasets to the Counseling Dataset, which had the fewest samples and the most significant class imbalance. Despite these constraints, the Counseling Dataset remains highly valuable for its unique perspective on online mental health conversations, particularly in seeking expert advice.

\textbf{Annotation Process.} The GTC already contains a binary trauma variable that psychologists have annotated according to the APA definition of trauma.
%with two other psychologists labeling a subset of $n = 200$ samples for annotator agreement (Fleiss' $\kappa = 0.84$) \citep{schirmer2023uncovering}. 
For the PTSD and Counseling datasets, 1,200 instances each were annotated by crowdworkers. We used the Portable Text Annotation Tool (Potato; \citealp{pei2022potato}) to set up an annotation interface for crowdworkers using Prolific as a recruitment platform for annotators. Each instance was labeled by three annotators, and all annotators received an hourly reimbursement of approximately 12 US\$. The crowdworkers were provided detailed instructions, the APA definition of a traumatic event, and three examples. Both the Prolific pre-screening and the instructions contained a trigger warning, ensuring that participants were free to pause or stop the study at any time (\Cref{sec:appendix}, \Cref{fig:annotator_instructions}). Annotators were based in either the US or the UK and fulfilled English language requirements.

We conducted a pilot study comparing single-choice and span annotation setups, where participants highlighted traumatic events in the text. The final annotation task used the span setup to ensure accurate detection (\Cref{sec:appendix}, \Cref{fig:annotator_interface}). Annotations were quality-checked, resulting in the removal of two annotator entries who labeled an unlikely number of samples as trauma, without affecting the total sample count (e.g., 1,200).
For the Incel dataset, we only labeled 300 instances since it serves as a control test set. To ensure quality, two researchers with psychology degrees annotated a subset of 200 instances from each dataset and resolved disagreements through discussion (Cohen's $\kappa = .82$).

\textbf{Annotator Agreement.} To assess annotator consistency, we report Krippendorff's $\alpha$ for agreement among crowdworkers and provide Binary F1 scores to measure agreement between the crowdworker majority vote and the expert vote, with the latter serving as the 'true' reference (Table \ref{tab:dataset_overview}). Both agreements were best for the Counseling Dataset. %, although this is potentially due to the small number of positive samples in the dataset. 
All agreement scores indicate at least moderate agreement \citep{krippendorff2018content}. Despite variability, our primary focus is on the accuracy of labels from majority voting. The moderate F1 scores indicate that majority votes are reliable labels, supporting the robustness of our annotation process. Given the subjective nature of interpreting trauma-related constructs, some disagreement is expected, similar to lower agreement seen in tasks like hate speech detection \citep{li2024hot}. This level of agreement, while not perfect, provides a solid foundation for the study.

%Overall, the results demonstrate that while the level of agreement among crowdworkers varies across datasets, the alignment between crowdworker majority votes and expert annotations remains relatively high, particularly in the Counseling Dataset.

\section{Methods}

\begin{table*}[t]
\centering
\small
\begin{tabular}{lccccc}
\toprule
\textbf{Model} & \textbf{Complexity} & \textbf{Interpretability} & \textbf{Hyperparameters} & \textbf{Scalability} & \textbf{Prediction} \\ \midrule
BoW-Naive-Bayes & Low & High & \begin{tabular}[c]{@{}c@{}} binary, \\ smoothing param.  $\alpha$\end{tabular} & High & After training \\ \hline
N-Gram Logistic Regression & Low & Medium & \begin{tabular}[c]{@{}c@{}}TF-IDF, \\ n-grams\end{tabular} & High & After training \\ \hline
TF-IDF Fully-Connected NN & Medium & Medium & \begin{tabular}[c]{@{}c@{}}Hidden layers, \\ layer width\end{tabular} & Medium & After training \\ \hline
BERT-based Models & High & Low & \begin{tabular}[c]{@{}c@{}}Learning rate, \\ layers, heads\end{tabular} & Low & \begin{tabular}[c]{@{}c@{}}One-shot or \\ after fine-tuning\end{tabular} \\ \hline
Black-box API (GPT-3.5/4) & High & Low & \begin{tabular}[c]{@{}c@{}}Prompt template, \\ API settings\end{tabular} & Low & \begin{tabular}[c]{@{}c@{}}One-shot or \\ after fine-tuning\end{tabular} \\ \bottomrule
\end{tabular}
\caption{Model Categorization According to General Suitability Criteria}
\label{tab:models}
\end{table*}

\subsection{Models and Hyperparameters}
In this work, we implement five sequence classification models for natural language inputs. The suitability of the these models for trauma detection in different contexts is defined by criteria such as complexity, interpretability, hyperparameter optimization, and scalability. To help in understanding the trade-offs and strengths of each approach, we provide an overview of the models considered in Table~\ref{tab:models}. The hyperparameters given are optimized with a hyperparameter optimization framework.

\noindent\textbf{BoW-Naive-Bayes Model.} The simplest model is obtained by fitting a Naive-Bayes model on the word counts in both classes. Let $\mathbf{t} = \lbrack t_1, t_2, \ldots, t_N \rbrack$ be an input sequence. 
We model the log-odds by combining two key components. First, we calculate the prior odds, which is the log of the initial ratio of the probabilities of the two categories. Second, we add the word-specific weights, which are summed over all elements in the input sequence. Each weight represents the log of the ratio of the probabilities of that element occurring in each category. 

%We model the log-odds as
%\begin{align}
%    \log \frac{p(y=1|\vt)}{p(y=0|\vt)} &=\\
%    \log \frac{p(y=1)}{p(y=0)} &+ \sum_{t_i \in \vt} \underset{w(t_i)}%{\underbrace{\log \frac{p(t_i|y=1)}{p(t_i|y=0)}}}.
%\end{align}

We obtain the weight of a term by counting its occurrences in documents from both classes and applying Laplace smoothing with a specified hyperparameter $\alpha$. The main advantage of this linear model is its interpretability due to the individual weights of each token that are explicitly computed.

%We obtain $w(\tau) = \frac{(\# \text{occ. of}~\tau~\text{in class 1})+\alpha}{(\# %\text{occ. of}~\tau~\text{in class 0}) +\alpha}$ by counting word frequencies in both %classes. We use Laplace smoothing with hyperparamter $\alpha$. 

\noindent \textbf{N-Gram Logistic Regression Model.} We compute $n$-grams for the datasets and fit a logistic regression model on the TF-IDF represenation of the $n$-Grams, where $n$ is $[1,2,3]$. % We use $n$ of up to three and consider the number of words $n$ to form the $n$-grams as a hyperparameter.

\noindent \textbf{TF-IDF Fully-Connected Model.} Furthermore, we compute TF-IDF vectors for the samples and train a fully connected neural network using this representation as an input. We use either one or two hidden layers, with the number of hidden layers and their width as a hyperparameter.

\noindent \textbf{BERT-based Models.} We train the popular encoder-only transformer models BERT \cite{devlin_et_al_2019} and RoBERTa \cite{liu2019roberta}. We experiment with both pretrained and non-pretrained versions of these models. We find that the pretrained models yield superior performance, which is why we restrict our analysis to these models for the main paper. We use the learning rate, number of layers, and number of heads as hyperparameters.

\noindent \textbf{Black-box API models (GPT-3.5/GPT-4).} We use a prompt template to access publicly available foundation model APIs for GPT-3.5 and GPT-4 \cite{achiam2023gpt}. We rephrase the classification tasks as a sequence completion tasks by using prompt template, which instructs the model to either output ``0'' or ``1'', and apply basic prompt engineering, including a task definition, the trauma definition, and labeling instructions (see Appendix \ref{sec:appx-prompt}). We use the top token log-probabilities returned by the API to compute class log-odds, which can be used to compute calibration measures and ROC curves.

%For the general 

%as 
%\begin{align}
%    \frac{p(y=1)}{p(y=0)} \approx \exp\left[\log(p(\text{``1''}))-%\log(p(\text{``0''}))\right]
%\end{align}

\subsection{Explainable AI Methods}
We use explainable AI approaches to gather insights on how trauma is described and recognized across different domains. Feature-based explanations allow us to gain insights into the importance of individual input features, i.e., tokens. We chose model-agnostic approaches that treat the predictive model as a black-box function and can be applied to any model (SHAP values) and model-specific, mechanistic approaches that are only applicable to specific models but can more faithfully describe the output of certain model classes.
Additionally, concept-based explanations allow us to move beyond individual feature attributions to a higher level of abstraction, and help us identify interpretable concepts that are crucial for trauma detection without requiring extensive supervision. These methods collectively enhance our ability to interpret model predictions and validate their reliability.

%\textbf{Linear Weight as Explanations.} The Na\"ive Bayes and the Logistic regression model fall into the class of inherently interpretable white-box models \cite{molnar2019}. For these models explanations can be obtained by simply inspecting their learned linear classification weights. For the more complex models, however, different forms of explanations are required.

\noindent \textbf{SHAP Explanations.} Shapley values originate from game theory and have been proposed to compute the contribution of individual features to the output of a non-linear function. They are a form of feature attribution explanation that assigns each input token a numerical score. The score corresponds to the average contribution to the output obtained when this feature is added. We compute SHAP values using an efficient sampling-based algorithm with the implementation of \citet{lundberg2017unified}.

\noindent\textbf{SLALOM Explanations.} \citet{leemann2024attention} have shown that single attribution scores cannot fully describe the inner workings of modern transformer language models. The authors propose SLALOM, a model to assess the role of input tokens along two dimensions: A \textit{token value} score, describes the effect each token has on its own, while the \textit{token importance} describes how much weight is placed on each token when tokens are concatenated to sequences. While SLALOM can be used to approximate any model's behavior in principle, it is particularly suited for transformer models, like the BERT and RoBERTa models used in this work. 

\noindent \textbf{Concept-based Explanations.} Concept-based explanations have been proposed as an alternative to feature-wise explanations. They do not reason over individual input features (tokens, pixels, etc.) but instead use a higher level of abstraction \cite{Kim2018interpretabilityTCAV, koh2020concept}. However, it is difficult to discover meaningful concepts from the data without supervision \cite{leemann2023post}. In case no concept annotations are present in the data, they identify clusters in a model's latent space that best describe a model's decision. In this work, we turn to Completeness-Aware Concept-Based Explanations \cite{yeh2019completeness}, which are one of the few conceptual explanation techniques that are applicable to textual inputs and do not require supervision in terms of the data. The concepts are represented as a set of salient examples, i.e., sample snippets that most strongly exhibit the discovered concept.

In this study, we focus on the RoBERTa architectures for concept-based text classification, which proved reliable across all datasets. We use the logit outputs of this model to obtain SHAP and SLALOM explanations and use the latent representation before the classification head as the latent space where the concept vectors are identified. Details on explanation approaches and their hyperparameters are provided in \Cref{sec:appxexplanations}. %We used our finetuned RoBERTa base model for each dataset to extract contextual embeddings from text, which were fed into ConceptBERT to identify and use concept vectors. 
%ConceptBERT integrated these embeddings, with concept vectors as trainable parameters for identifying relevant concepts.

\section{Model Performance Results}

\newcommand{\wstd}[2]{#1 \small{$\pm$ #2}}
\newcommand{\bwstd}[2]{\textbf{#1} \small{$\pm$ #2}}

\begin{table*}\small
     \setlength{\tabcolsep}{4pt}
    \centering
    \begin{tabular}{rcccccc}
    \toprule
    Dataset & \multicolumn{2}{c}{GTC} & \multicolumn{2}{c}{PTSD} &
    \multicolumn{2}{c}{Counseling}  \\
    \cmidrule(r){1-1} \cmidrule(lr){2-3}
    \cmidrule(lr){4-5}\cmidrule(l){6-7}
    LM & F1 (bin.) & AU-ROC & F1 (bin.) & AU-ROC & F1 (bin.) & AU-ROC \\
     \midrule
NaiveBayes-BoW &\wstd{0.53}{0.09}  & \wstd{0.82}{0.09} & \wstd{0.56}{0.04}  & \wstd{0.70}{0.02} & \wstd{0.17}{0.01}  & \wstd{0.70}{0.02} \\
NGramLogisticRegression & \wstd{0.51}{0.10}  & \wstd{0.83}{0.09}  &  \wstd{0.58}{0.02}  & \wstd{0.70}{0.02} & \wstd{0.15}{0.05}  & \wstd{0.79}{0.01} \\
FeedForwardModel & \wstd{0.52}{0.10}  & \wstd{0.84}{0.09} & \wstd{0.52}{0.05}  & \wstd{0.74}{0.01} & \wstd{0.03}{0.03}  & \wstd{0.78}{0.01}  \\
%BERTmodel & \wstd{0.904}{0.002}  & \wstd{0.613}{0.016}  & \wstd{0.899}{0.013} \\
%RoBERTamodel & \wstd{0.909}{0.002}  & \wstd{0.634}{0.022}  & \wstd{0.939}{0.002} \\
BERT (finetuned) & \wstd{0.71}{0.01}  & \wstd{0.96}{0.00}  &\wstd{0.66}{0.02}  & \wstd{0.80}{0.01} & \bwstd{0.35}{0.05}  & \bwstd{0.91}{0.01} \\
RoBERTa (finetuned) & \bwstd{0.74}{0.01}  & \bwstd{0.97}{0.00} & \bwstd{0.71}{0.01}  & \bwstd{0.83}{0.01} & \wstd{0.18}{0.09}  & \wstd{0.88}{0.02} \\
OpenAI GPT-4 & 0.64 & 0.94 & 0.69 & 0.82 & \textbf{0.36}  & 0.85\\
     \bottomrule
    \end{tabular}
    \caption{Classification performance of the language models used in this work. We report Binary F1-Scores, and Area under the Receiver-Operator Curve (``AU-ROC''). We report standard errors over cross-validation with 5 runs for all models but the Black-box API models, where computation costs are prohibitive. \label{tab:modelperf}}
\end{table*}

\paragraph{Classification Performance} We fit all the models to the respective datasets after performing hyperparameter optimization (cf. \Cref{sec:appx-prompt}) and report their performance metrics in \Cref{tab:modelperf}.
The evaluation across GTC, PTSD, and Counseling datasets shows clear trends. Transformer-based models, especially fine-tuned BERT and RoBERTa, significantly outperform traditional models and feedforward neural networks. The Naive-Bayes-BoW and NGram Logistic Regression models show moderate performance but lag behind due to their simpler architectures. The feedforward model performs reasonably well but is outclassed by transformer models. Fine-tuned BERT and RoBERTa exhibit substantial improvements in all metrics, with RoBERTa achieving the highest F1 scores in the GTC dataset ($F1 = .74$) and the PTSD dataset ($F1 = .71$), highlighting its effective language comprehension capabilities. To control for dataset size effects, we ran an additional experiment using 1,000 randomly selected GTC samples in the training set to match the size of other datasets. The performance remained consistent, indicating that our findings on smaller datasets likely extend to larger ones (\Cref{sec:appendix}, \Cref{tab:gtc-small}).

OpenAI's GPT-4 also performs particularly well on the PTSD and Counseling datasets and even outperforms BERT in the F1 metric on Counseling, showcasing its strong generalization abilities despite not being further fine-tuned and relying on a single prompt for these tasks. Interestingly, all models perform reasonably well, which may be attributed to the specific task of trauma event detection. However, the Counseling dataset proved more challenging due to its very imbalanced class distribution and the presence of very few trauma event samples. This is reflected GPT-4 F1 score of $.36$, which was the highest for this dataset but still indicates the difficulty of the task. 
RoBERTa achieves strong performance metrics overall, highlighting the impact of architectural improvements and extensive training on larger datasets, though it does not outperform BERT on the Counseling dataset.

\begin{figure}[htbp]
    \centering
    \includegraphics[width=\columnwidth]{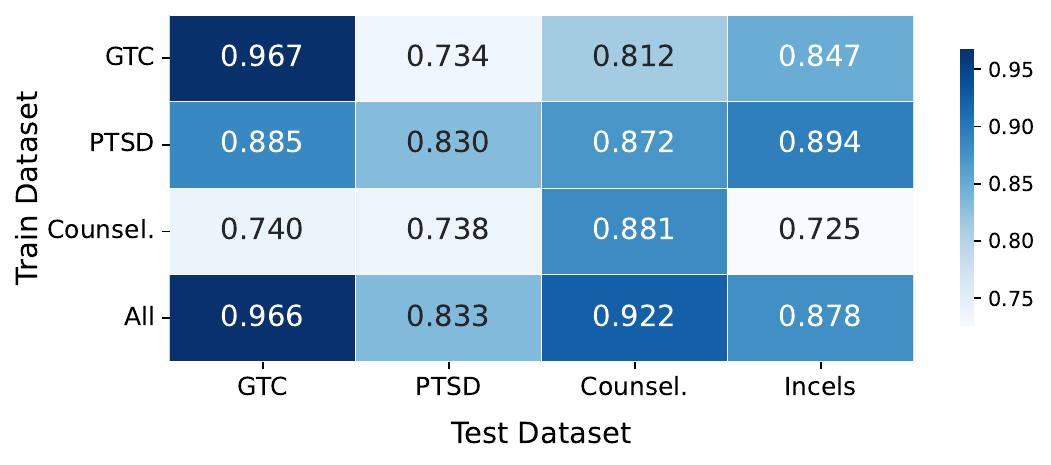}
    \caption{Cross-domain performance (AUC-ROC) when a RoBERTa model is trained on one dataset and tested on other datasets. }
    \label{fig:heatmap}
\end{figure}

\paragraph{Cross-Domain Performance}
\Cref{fig:heatmap} presents the cross-domain results of RoBERTa models fine-tuned on one dataset and evaluated on other datasets, using the AUC-ROC metric (cf., \Cref{sec:appendix}, \Cref{tab:crosstest}). Models trained on the GTC dataset showed the highest generalizability, performing well across all test sets. Those trained on the PTSD dataset excelled on their own test set and performed strongly on others. Models trained on the Counseling dataset achieved top performance on their own set but did less well on others. The model trained on all combined datasets showed robust and consistent performance across all test sets, maintaining high accuracy and reliability.
Despite differences in trauma types across datasets, significant overlaps contribute to strong cross-testing results. For example, both the GTC and PTSD datasets include trauma related to death, acute stress reactions, and physical violence, aiding models' cross-dataset performance. However, the GTC dataset's unique military component may cause some performance differences. Overall, high cross-domain performance suggests that shared trauma themes enable effective generalization across different contexts.

The results show that the RoBERTa model fine-tuned on the PTSD dataset has the best generalizability across different datasets, with models trained on the full data also performing well. Given the diversity of traumatic events across datasets, this result suggests the trauma features in the PTSD dataset are broadly applicable for learning a general event type, rather than causing models to pick up on only keywords. Counseling-trained models perform well on their own dataset but do not generalize as effectively. Performance on the Incel dataset indicates all models effectively differentiate trauma-related vocabulary from control data.

\paragraph{SHAP Explanations}
To understand how the models attribute feature importance to the trauma label, we calculated SHAP values for some samples from all datasets, focusing on comparing RoBERTa and GPT-4 due to their high performances and the interesting differences in how these language models classify trauma. While most classifications aligned (see \Cref{fig:combined_shap_values} in Appendix \ref{sec:appendix}), we found that, in several instances, GPT-4 provided more non-trauma attributions for certain features compared to RoBERTa.

\begin{figure}[ht]
    \centering
    \begin{subfigure}[b]{0.45\textwidth}
        \centering
        \includegraphics[width=\textwidth]{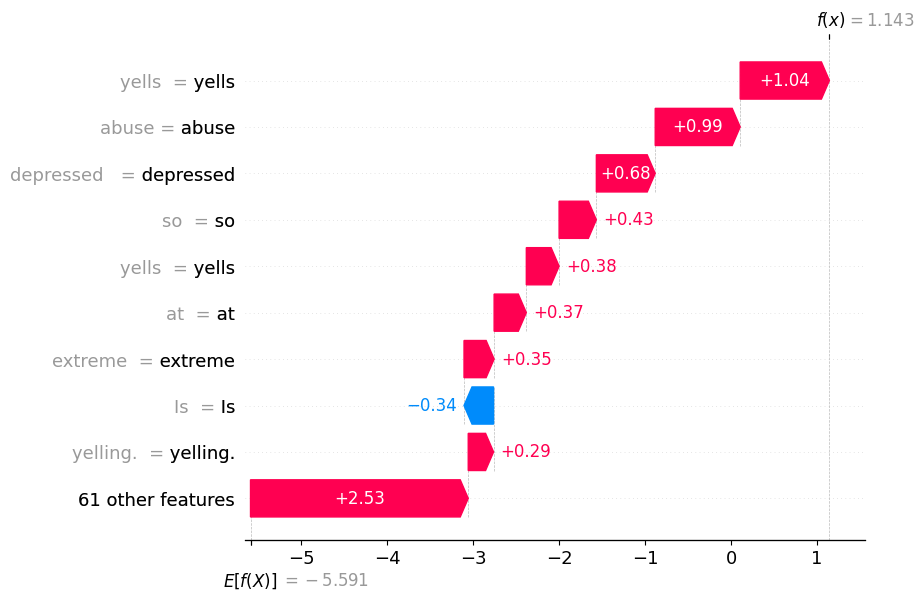}
        \caption{RoBERTa}
        \label{fig:first_shap}
    \end{subfigure}
    \hfill
    \begin{subfigure}[b]{0.45\textwidth}
        \centering
        \includegraphics[width=\textwidth]{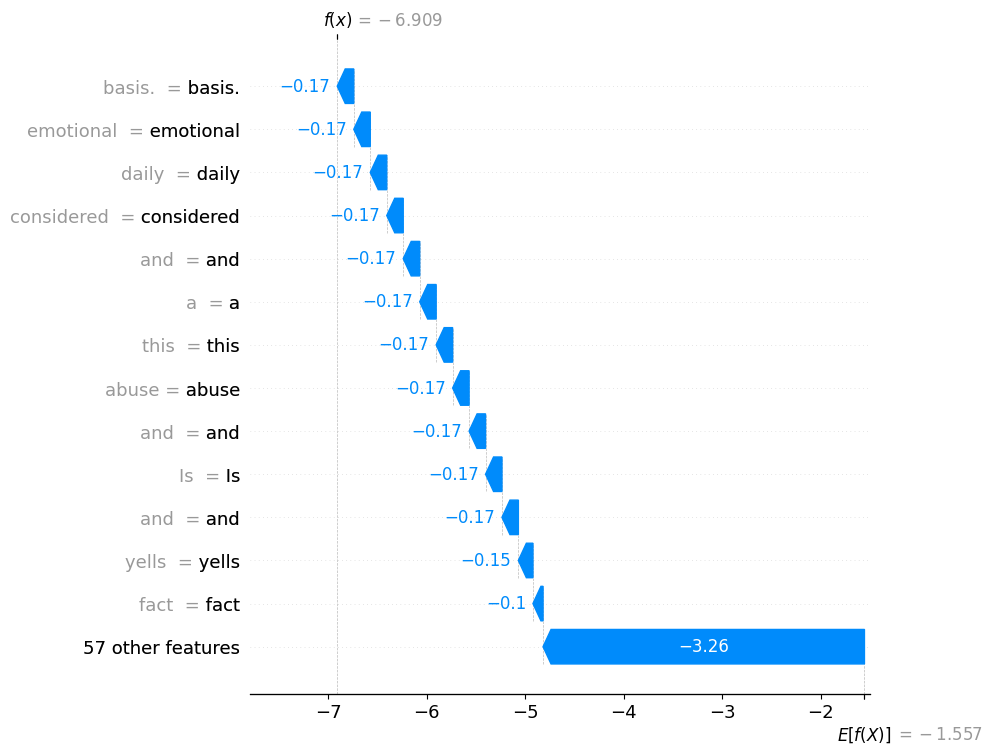}
        \caption{GPT-4}
        \label{fig:second_shap}
    \end{subfigure}
    \caption{SHAP values for an instance from the \textbf{Counseling Dataset}: \enquote{My dad doesn't like the fact that I'm a boy. He yells at me daily because of it and he tells me I'm extreme and over dramatic. I get so depressed because of my dad's yelling. He keeps asking me why I can't just be happy the way I am and yells at me on a daily basis. Is this considered emotional abuse?}}
    \label{fig:disagreement_instance_counseling}
\end{figure}

Figure \ref{fig:disagreement_instance_counseling} shows a counseling dataset example where RoBERTa and GPT-4 disagree. RoBERTa assigns high relevance to words like \emph{yells}, \emph{abuse}, and \emph{depressed}, while GPT-4 does not, possibly due to the forum user's uncertainty about defining abuse. This discrepancy may stem from GPT-4's closer adherence to the APA definition of trauma, with less variation and personal bias than human annotators, who may classify events based on their own experiences and interpretations.

These findings, though based on exemplary instances, highlight the challenge of detecting mental abuse. RoBERTa may rely more on specific keywords related to abuse, whereas GPT-4 seems to consider contextual nuances. Human annotators might interpret such incidents as traumatic based on subjective judgment and empathy, while GPT-4, adhering strictly to the APA definition of trauma, did not classify these incidents as trauma.

\section{Characteristics of Trauma Across Domains}

\begin{figure}[htbp]
    \centering
    \includegraphics[width=\columnwidth]{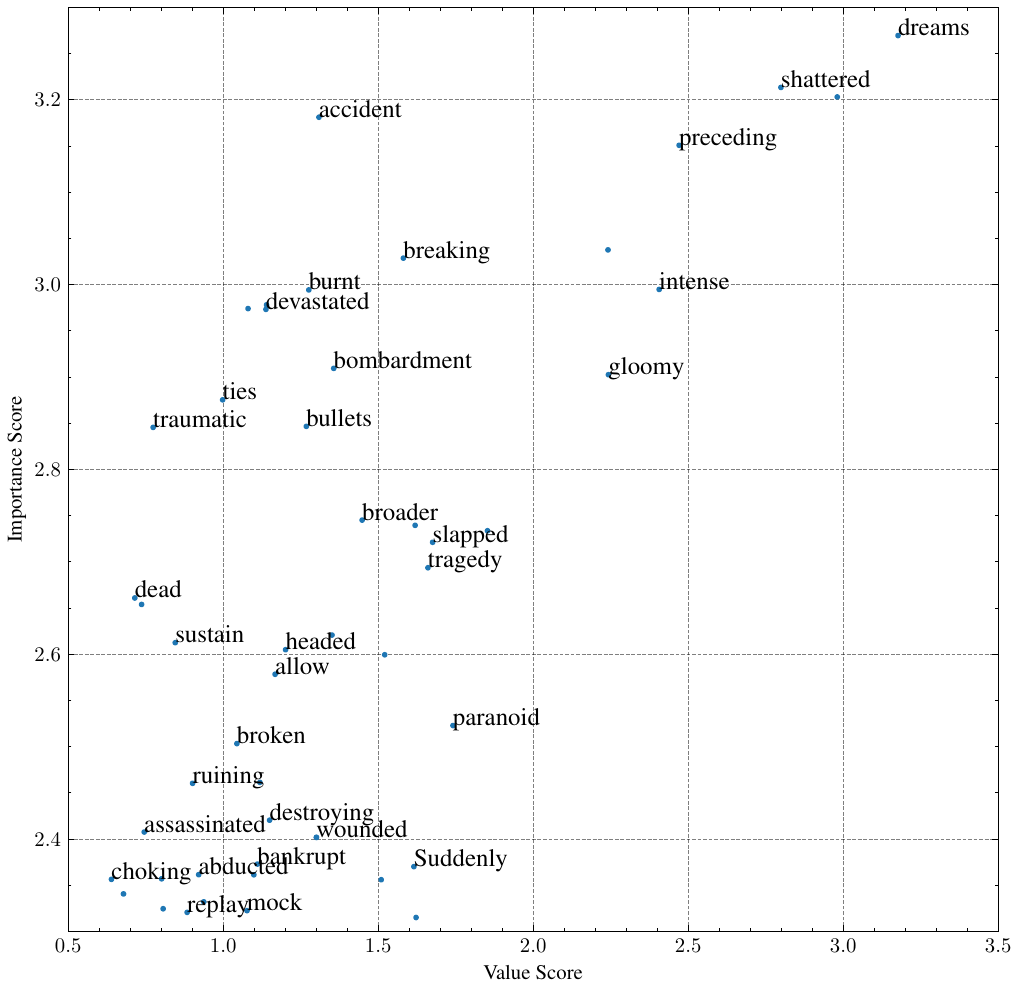}
    \caption{SLALOM feature importance scores based on the full dataset and the RoBERTa model.}
    \label{fig:slalom}
\end{figure}

\paragraph{Feature Characteristics with SLALOM} The SLALOM feature importance scores from all datasets focus on the highest value features for trauma classification. Features like \emph{dream} and \emph{shattered}, in the top right corner, contribute most to the trauma classification. For clarity, overlapping features were excluded (blue dots remain in the figure) (\Cref{fig:slalom}).

Notable feature variability includes war-related vocabulary (e.g., \emph{bombardment}, \emph{bullets}) likely from genocide-related data, and more generalizable words (e.g., \emph{dreams}, \emph{accident}, \emph{dead}) applicable across domains. Amplifying words like \emph{intense}, \emph{suddenly}, and \emph{gloomy} also appear, fitting traumatic contexts without specific events.

Groups of thematically related words are evident: \emph{dead} and \emph{assassinated} represent death, \emph{wounded}, \emph{choking}, and \emph{slapped} indicate physical injury and violence, and \emph{dreams}, \emph{shattered}, and \emph{replay} are associated with trauma's psychological impact.

\newcommand{\boxwidth}[0]{5.5cm}
\newcommand{\fixheightbox}[1]{
  \begin{minipage}[t][8cm][t]{\textwidth}
   \parbox{\boxwidth}{\setlength{\baselineskip}{0.9\baselineskip}
   #1}
  \end{minipage}
}

\newcommand{\highlight}[1]{\colorbox{yellow}{#1}}

\begin{figure}[htbp]
\centering
\begin{minipage}[t]{0.5\textwidth}
    \centering
    \begin{subfigure}[t]{0.48\textwidth}
\centering
    \scalebox{0.6}{
    \begin{tikzpicture}
    % Define the box
    \node[draw=brown, thick, rounded corners, text width=\boxwidth, minimum height = 9.5cm, inner sep=10pt] (box2) 
    {\centering
        % Heading
        \textbf{GTC: Concept 4} \\
        % Draw a horizontal line
        \tikz \draw[brown, thick] (0,0) -- (\linewidth,0);\vspace{1em}
        \fixheightbox{
and when he \highlight{attacked} me\\
 chief, was \highlight{very cruel}\\
 I was \highlight{punished} that way\\
  He \highlight{pressed} me against\\
 His disappearance was very \highlight{painful}\\
 Bou Meng was \highlight{tortured} for\\
 who \highlight{tortured} me was Si\\
 so I had him buried\\
, they stopped \highlight{beating} me\\
 who tore the child away\\
 all the \highlight{beatings} that\\
 They started \highlight{beating} me,\\
 task of killing people.
        }
    };
\end{tikzpicture}}
\caption{torture, abuse\\~}
\end{subfigure}
\end{minipage}\hfill
\begin{minipage}[t]{0.5\textwidth}
    \centering
    \begin{subfigure}[t]{0.48\textwidth}
\centering
    \scalebox{0.6}{
    \begin{tikzpicture}
    % Define the box
    \node[draw=brown, thick, rounded corners, text width=\boxwidth, minimum height = 9.5cm, inner sep=10pt] (box2) 
    {\centering
        % Heading
        \textbf{PTSD: Concept 9} \\
        % Draw a horizontal line
        \tikz \draw[brown, thick] (0,0) -- (\linewidth,0);\vspace{1em}
        \fixheightbox{
 extremely frequent \highlight{flashbacks} the \\
 me bad. The \highlight{flashbacks}\\
 When I was \highlight{molested}\\
 have vivid \highlight{flashbacks}. All\\
 about \highlight{flashbacks}.
I\\
 after i was \highlight{sexually assaulted}\\
 young child, was sexually\\
 having nightmares and \highlight{flashbacks}\\
 \highlight{repressed memories} are a\\
 because I have \highlight{flashbacks} several\\
 always thought the \highlight{memories}\\
 of scolding via email\\
 like I was \highlight{abused}.
        }
    };
\end{tikzpicture}}
\caption{flashbacks, abuse}
\end{subfigure}
\end{minipage}\hfill
\begin{minipage}[t]{0.5\textwidth}
    \centering
    \begin{subfigure}[t]{0.48\linewidth}
\centering
    \scalebox{0.6}{
    \begin{tikzpicture}
    % Define the box
    \node[draw=brown, thick, rounded corners, text width=\boxwidth, minimum height = 9.5cm, inner sep=10pt] (box2) 
    {\centering
        % Heading
        \textbf{Counseling: Concept 9} \\
        % Draw a horizontal line
        \tikz \draw[brown, thick] (0,0) -- (\linewidth,0);\vspace{1em}
        \fixheightbox{
I was violently \highlight{raped} by\\
 got \highlight{pregnant} by my boyfriend\\
 my \highlight{baby} mother. She\\
 my children's father left\\
I saw my mother cheating\\
I was \highlight{raped} by multiple\\
My girlfriend was \highlight{abused} as\\
 I got \highlight{raped} by my\\
 I just lost my mom\\
 teenager. My entire family\\
I was \highlight{raped} repeatedly when\\
My grandma and brother both\\
 parents injured my brother,\\
, my husband mentally abused\\
My mother has Alzheimer's
        }
    };
\end{tikzpicture}}
\caption{rape, pregnancy}
\end{subfigure}
\end{minipage}
\caption{Trauma-related concepts found in the three datasets (Most salient examples, RoBERTa Model). For more examples see \Cref{sec:appendix}.
\label{fig:concepts_main}}
\end{figure}

\paragraph{Conceptual Explanations} For each dataset, we assessed conceptual explanations to detect context-specific trauma concepts. We select the concepts that have the highest number of traumatic instances in the neighborhood closely associated with the corresponding concept (\Cref{fig:concepts_main}).

In the genocide dataset, concepts related to killings, death, and severe injuries were prominent, reflecting the extreme nature of the content. In contrast, the PTSD and counseling datasets, which address more everyday trauma, contained more references to domestic violence and abuse. The smaller size of the counseling dataset made it challenging to identify unique concepts without overlap.

Across all contexts, death and sexual violence were prevalent. In the genocide dataset, these were depicted through killings and executions, whereas in other datasets, they were associated with grief, loss, and suicide. Sexual violence, particularly rape, consistently appeared as a common source of PTSD, which is consistent with the psychological literature \cite{atwoli2015epidemiology}.

\section{Conclusion}

Traumatic events shape millions of lives. Computational tools to recognize these events can help third parties provide support. However, their diversity makes classification challenging. This paper introduces a new dataset for recognizing traumatic events and analyzes (i) NLP models' performance, (ii) their generalizability across domains, and (iii) if they learn general trauma features using XAI techniques. We show that transformer-based models offer strong performance and generalization, though simpler models still perform well in-domain. However, zero-shot performance by GPT-4 lags behind fine-tuned models.
%
%We show that transformer-based models, particularly fine-tuned BERT and RoBERTa, outperform traditional models and neural networks, although simpler models achieve very close scores. Models trained on diverse datasets, like the combined dataset model, perform well across various contexts, emphasizing the importance of exposure to varied data. Despite unique aspects in datasets like GTC's military component, shared trauma themes enable strong cross-domain generalizability. Notably, large language models like GPT-4 do not necessarily improve performance on this task without fine-tuning and domain-specific adjustments.
%
Our analysis shows that while certain features of trauma are context-specific, there are also universal elements across different experiences. However, certain types of traumatic events---notably mental abuse---are particularly challenging to classify due to their less defined nature and greater variability, highlighting the need for clear definitions and enhanced model performance. % This highlights the importance of clearly defining and conceptualizing traumatic events to enhance model detection and classification accuracy (ADD CITATION).

\section{Limitations}
%Our study has several limitations. 
The different contexts of the datasets and label imbalance, especially in the Counseling dataset, affect the cross-testing results and overall model performance in trauma detection. Label imbalance is particularly challenging because models may become biased towards the more frequent non-trauma events, leading to poorer performance in detecting the less common trauma events. It is normal to have a smaller number of trauma event samples, making it harder for models to learn and accurately identify these underrepresented cases. However, given that the primary goal of this study is to demonstrate cross-domain transferability on realistic data, it is essential to use datasets with an expected and realistic class distribution.

Technical limitations include the summative nature of the explanations, which only provide high-level insights into the different natures of trauma across domains. Additionally, sampling-based explanations such as SLALOM and SHAP are only approximations of the true model behavior, and their fidelity can be increased with more samples, though this incurs higher computational costs.

Another limitation is that people discuss traumatic events differently depending on the context, which might limit the comparability of the datasets used in this study. Conversations with mental health professionals often use clinical terms, focusing on symptoms, triggers, and coping mechanisms \citep{tong2019talking}, while online forums blend informal and semi-formal language where anonymity allows for candid sharing, but responses may vary in depth and understanding \citep{lahnala2021exploring, stana2017battling}. This contrasts with court testimonies, which require precise, factual language focused on specific events and details for legal documentation \citep{ciorciari2011trauma, schirmer2023talking}.

We chose the span annotation method, where annotators select the text indicating a traumatic event, because pilot experiments showed it improved performance by focusing attention on specific events rather than a simple "yes" or "no" decision. Although this was a design choice and not a central research question, analyzing these spans could offer insights into annotation quality and inform future training. Investigating the detection of specific traumatic event spans rather than general segments is a promising direction for future research.

Finally, our analysis partially relies on social media data. This type of data provides vast, real-time insights into public mental health trends but can be noisy and less reliable. It would be important for future studies to replicate our results with clinical data to ensure the findings' robustness and applicability in medical settings.

\section*{Ethics Statement}

Our data processing procedures did not involve any handling of private information. No user names were obtained at any point of the data collection process.
%To promote responsible usage and prevent potential misuse, we make our annotated data available upon request, considering the data's violent nature and the need to address concerns about circumventing hate speech detection.
The human annotators were informed of and aware of the potentially violent content before the annotation process, with the ability to decline annotation at any time. The same is true for crowdworkers, who were presented several trigger warnings throughout the process. Both human coders were given the chance to discuss any distressing material encountered during annotation. As discussions on the potential trauma or adverse effects experienced by annotators while dealing with distressing material become more prevalent \citep{kennedy2022introducing}, we have proactively provided annotators with a recommended written guide designed to aid in identifying changes in cognition and minimizing emotional risks associated with the annotation process.

\bibliography{acl_latex}

%\balance

%\pagebreak

\appendix

\section{Appendix}
\label{sec:appendix}

\subsection{Implementation Details: Explanation Methods}
\label{sec:appxexplanations}
In this section, we give more details on how we computed the explanations shown in this paper.

\textbf{SHAP Values.} To obtain SHAP values, we use the official \texttt{shap}\footnote{https://github.com/shap/shap} package. We use the TextExplainer class.

\textbf{SLALOM.} We use the SGD algorithm proposed in \citet{leemann2024attention} to estimate the SLALOM model on 100k background samples of length 2. We use all the tokens that appear in the samples from the datasets used and fit one global SLALOM model.

\textbf{Conceptual Explanations.} 
We use the completeness-aware loss proposed by \citet{yeh2019completeness} with snippets of length of 5 token as snippets for the algorithm. 
We trained with concept discovery module to discover $K=10$ concepts using the Adam optimizer at an initial learning rate of $1\times10^{-3}$, decaying to $5\times10^{-4}$ and $1\times10^{-4}$ in subsequent epochs. Training lasted 3 epochs with a batch size of 12. The model weights used were obtained from the best-performing model. We identified the 25 closest activations per concept. Evaluation on a separate test set involved dot products between latent representations and concept vectors, selecting the top activations.

\subsection{Implementation Details: Models}
\label{sec:appx-prompt}

We use the \texttt{optuna}\footnote{https://optuna.org/} framework for hyperparameter optimization with 50 steps for each model/dataset. We then train the models using different seeds and on five random data splits using the discovered hyperparameters. Through the optimization we obtain the parameters given in \Cref{tab:model_parameters}.

\textbf{Prompt Template.} We use the following prompt template to prompt the GPT models as the system prompt. 

\textit{"You are tasked with detecting trauma in text segments of transcripts of genocide tribunals. Specifically, detect instances that meet the APA’s definition of trauma. Psychological trauma, as defined by the APA, includes experiences of exposure to actual or threatened death, serious injury, or sexual violence, either directly encountered or witnessed. It also includes instances where individuals learn that the traumatic event(s) occurred to a close family member or friend. Label the text with '1' if there are indicators of trauma based on this definition, and '0' if there are no indicators of trauma. Note that trauma is rare and occurs in less than 20\% of the cases. Only answer with either '0' or '1'."
}

The samples are then passed as a user prompt.
\begin{table*}[ht]\small
    \centering
    \begin{tabular}{llll}
        \toprule
        & \multicolumn{3}{c}{Parameters}\\
        \textbf{Model} & \textbf{GTC} & \textbf{PTSD} & \textbf{Counseling} \\
        \midrule
        NaiveBayes-BoW & \begin{tabular}[c]{@{}l@{}}multiplicities: true \\ alpha: 1.01 \end{tabular} & \begin{tabular}[c]{@{}l@{}}multiplicities: true \\ alpha: 5.97 \end{tabular}& \begin{tabular}[c]{@{}l@{}}multiplicities: false \\ alpha: 1.01 \end{tabular} \\
        \midrule
        NGramLogisticRegression & \begin{tabular}[c]{@{}l@{}}n\_gram\_range: [1, 2] \\ C: 0.92 \\ penalty: l2 \end{tabular} &
        \begin{tabular}[c]{@{}l@{}}n\_gram\_range: [2, 3] \\ C: 0.0 \\ penalty: none \end{tabular} & \begin{tabular}[c]{@{}l@{}}n\_gram\_range: [1, 2] \\ C: 9.36 \\ penalty: l2 \end{tabular} \\
        \midrule
        FeedForwardModel & \begin{tabular}[c]{@{}l@{}}hidden\_dim1: 50 \\ hidden\_dim2: 80 \\ lr: 5.72e-05 \end{tabular}
        & \begin{tabular}[c]{@{}l@{}}hidden\_dim1: 50 \\ hidden\_dim2: none \\ lr: 1.79e-04 \end{tabular}
        & \begin{tabular}[c]{@{}l@{}}hidden\_dim1: 200 \\ hidden\_dim2: 50 \\ lr: 5.72e-05 \end{tabular} \\
        \midrule
        BERT (finetuned) & \begin{tabular}[c]{@{}l@{}}n\_layers: 5 \\ lr: 2.32e-05 \end{tabular}
        & \begin{tabular}[c]{@{}l@{}}n\_layers: 12 \\ lr: 1.10e-05 \end{tabular}
        & \begin{tabular}[c]{@{}l@{}}n\_layers: 6 \\ lr: 1.41e-05 \end{tabular} \\
        \midrule
        RoBERTa (finetuned) & \begin{tabular}[c]{@{}l@{}}n\_layers: 12 \\ lr: 2.04e-06 \end{tabular} & \begin{tabular}[c]{@{}l@{}}n\_layers: 7 \\ lr: 6.43e-06 \end{tabular} & \begin{tabular}[c]{@{}l@{}}n\_layers: 4 \\ lr: 9.54e-05 \end{tabular} \\
        \midrule
        OpenAI & \begin{tabular}[c]{@{}l@{}}target\_model: gpt-4-turbo \end{tabular} & \begin{tabular}[c]{@{}l@{}} target\_model: gpt-4-turbo \end{tabular}&  \begin{tabular}[c]{@{}l@{}} target\_model: gpt-4-turbo \end{tabular} \\
        \bottomrule
    \end{tabular}
    \caption{Automatically selected hyperparameters for the different datasets}
    \label{tab:model_parameters}
\end{table*}

\subsection{Annotation Details}
\label{sec:app-instructions}

Participants were prescreened using Prolific based on self-reported English-language proficiency. We did not collect demographic data from the annotators as such data was not central to the questions our study is focused on and Prolific does not normally include this metadata.

\begin{figure*}[ht]
    \centering
    \includegraphics[width=\textwidth]{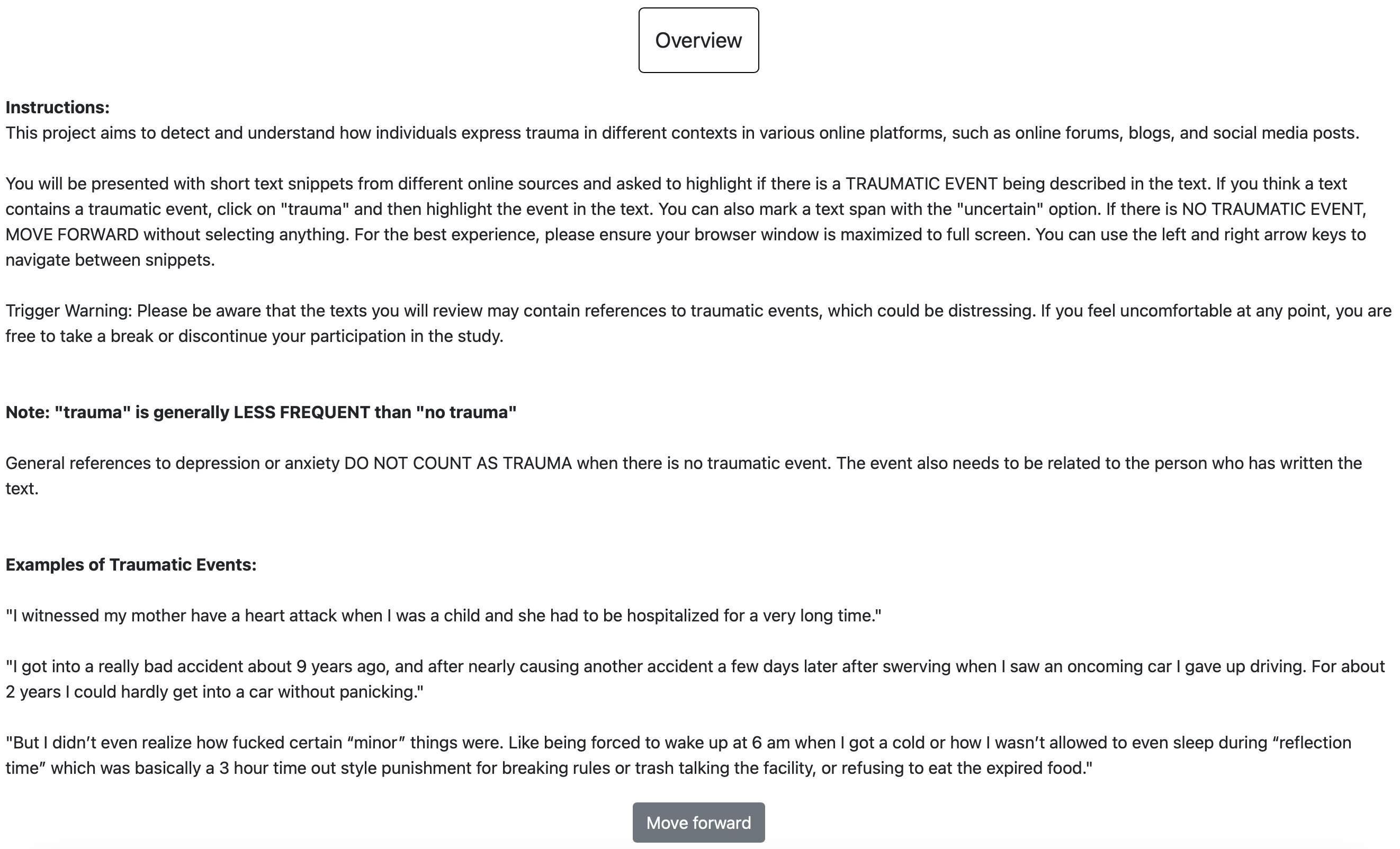}
    \caption{Instructions for Annotators. Note: We selected these examples because they were the most frequently mislabeled in the pilot, making them particularly relevant. Additionally, we kept the instruction page concise to avoid overwhelming the annotators, as excessive detail could deter them or lead to less careful reading.}
    \label{fig:annotator_instructions}
\end{figure*}

\begin{figure*}[ht]
    \centering
    \includegraphics[width=\textwidth]{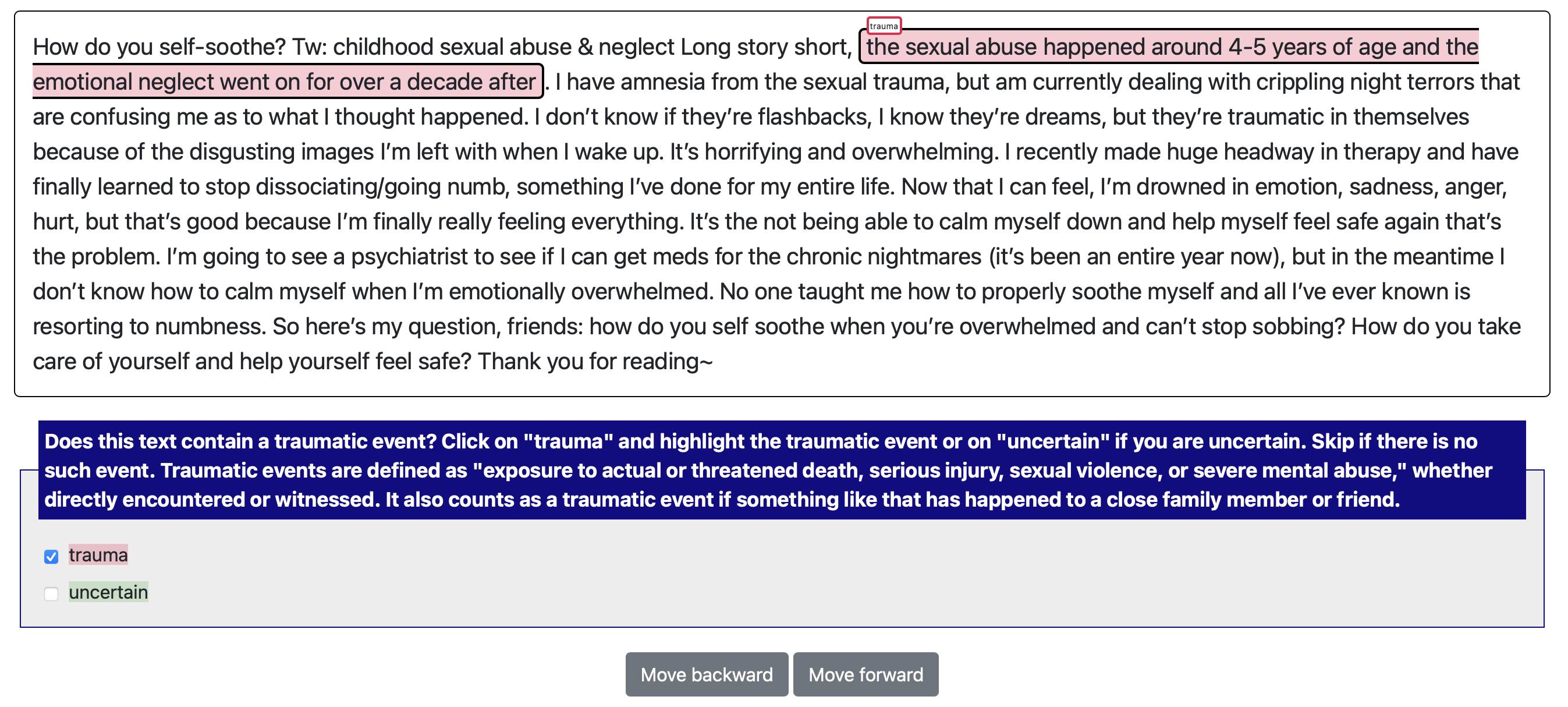}
    \caption{Interface of the Span Annotation Task.}
    \label{fig:annotator_interface}
\end{figure*}

\begin{table*}[ht]\small
     \setlength{\tabcolsep}{4pt}
    \centering
    \begin{tabular}{rcccc}
    \toprule
    Dataset & \multicolumn{3}{c}{Test Dataset}\\
    \cmidrule(r){1-1} \cmidrule(l){2-5}
    Train & GTC & PTSD & Counsel. & Incels\\
     \midrule
GTC & \bwstd{0.967}{0.000}  & \wstd{0.734}{0.005}  & \wstd{0.812}{0.020}  & \wstd{0.847}{0.003} \\
PTSD  & \wstd{0.885}{0.010}  & \wstd{0.830}{0.006}  & \wstd{0.872}{0.014}  & \bwstd{0.894}{0.010} \\
Counsel. & \wstd{0.740}{0.017}  & \wstd{0.738}{0.018}  & \wstd{0.881}{0.016}  & \wstd{0.725}{0.027} \\
All & \wstd{0.966}{0.001}  & \bwstd{0.833}{0.013}  & \bwstd{0.922}{0.012}  & \wstd{0.878}{0.005} \\
     \bottomrule
    \end{tabular}
    \caption{Cross-Testing models trained on one dataset on other datasets. Model: RoBERTa finetuned with AU-ROC metric\label{tab:crosstest}}
\end{table*}

\subsection{Metrics}
For completeness, we additionally report accuracy, recall, and precision for the trained models in \Cref{tab:metricsothers}.

\begin{table*}[htbp]
    \centering
    \begin{tabular}{lccc}
    \toprule
    Model & Accuracy & Precision & Recall \\
    \midrule
    \multicolumn{4}{c}{GTC}\\
    \midrule
    NaiveBayesBOWmodel & \wstd{0.84}{0.03}  & \wstd{0.44}{0.08}  & \wstd{0.69}{0.12} \\
NGramLogisticRegression & \wstd{0.88}{0.02}  & \wstd{0.60}{0.12}  & \wstd{0.44}{0.09} \\
FeedForwardModel & \wstd{0.88}{0.02}  & \wstd{0.60}{0.12}  & \wstd{0.46}{0.09} \\
BERTmodel & \wstd{0.88}{0.03}  & \wstd{0.58}{0.12}  & \wstd{0.46}{0.10} \\
RoBERTamodel & \wstd{0.91}{0.00}  & \wstd{0.70}{0.02}  & \wstd{0.59}{0.05} \\
BERTPretrainedmodel & \wstd{0.92}{0.00}  & \wstd{0.74}{0.03}  & \wstd{0.70}{0.04} \\
RoBERTaPretrainedmodel & \wstd{0.93}{0.00}  & \wstd{0.75}{0.03}  & \wstd{0.74}{0.04} \\
OpenAI GPT-4 & 0.91	& 0.68 & 0.61\\
\bottomrule
\multicolumn{4}{c}{PTSD}\\
    \midrule
    NaiveBayesBOWmodel & \wstd{0.69}{0.01}  & \wstd{0.63}{0.02}  & \wstd{0.52}{0.06} \\
NGramLogisticRegression & \wstd{0.68}{0.01}  & \wstd{0.62}{0.03}  & \wstd{0.54}{0.03} \\
FeedForwardModel & \wstd{0.70}{0.01}  & \wstd{0.71}{0.04}  & \wstd{0.42}{0.05} \\
BERTPretrainedmodel & \wstd{0.72}{0.01}  & \wstd{0.64}{0.01}  & \wstd{0.69}{0.06} \\
RoBERTaPretrainedmodel & \wstd{0.75}{0.01}  & \wstd{0.66}{0.02}  & \wstd{0.78}{0.04} \\
OpenAI GPT-4 & 0.69	& 0.58 & 0.84 \\
\bottomrule
\multicolumn{4}{c}{Counseling}\\
    \midrule
    NaiveBayesBOWmodel & \wstd{0.26}{0.01}  & \wstd{0.09}{0.01}  & \wstd{0.99}{0.01} \\
NGramLogisticRegression & \wstd{0.92}{0.01}  & \wstd{0.55}{0.17}  & \wstd{0.09}{0.03} \\
eedForwardModel & \wstd{0.92}{0.01}  & \wstd{0.10}{0.10}  & \wstd{0.02}{0.02} \\
BERTPretrainedmodel & \wstd{0.93}{0.01}  & \wstd{0.54}{0.04}  & \wstd{0.27}{0.05} \\
RoBERTaPretrainedmodel & \wstd{0.91}{0.01}  & \wstd{0.36}{0.19}  & \wstd{0.20}{0.12} \\
OpenAI GPT-4 & 0.91	& 0.42 & 0.31\\

\bottomrule
    \end{tabular}
    \caption{Additional model performance metrics. We see that the non-pretrained versions of BERT/RoBERTa do not perform on par with the pretrained ones on GTC. Therefore, we consider only the pretrained versions for the rest of the paper.}
    \label{tab:metricsothers}
\end{table*}

\begin{table*}[htbp]
    \centering
    \begin{tabular}{lcccc}
    \toprule
       & \multicolumn{2}{c}{GTC-1000} & \multicolumn{2}{c}{GTC-All}\\
    \cmidrule(r){1-1} \cmidrule(lr){2-3}  \cmidrule(lr){4-5}
    LM & F1 (bin.) & AU-ROC & F1 (bin.) & AU-ROC\\
    \midrule
FeedForwardModel & \wstd{0.38}{0.01}  & \wstd{0.86}{0.00} & \wstd{0.52}{0.10}  & \wstd{0.84}{0.09} \\
BERTPretrainedmodel & \wstd{0.61}{0.03}  & \wstd{0.93}{0.00} & \wstd{0.71}{0.01}  & \wstd{0.96}{0.00} \\
RoBERTaPretrainedmodel & \wstd{0.66}{0.03}  & \wstd{0.95}{0.00} & \wstd{0.74}{0.01}  & \wstd{0.97}{0.00} \\
\bottomrule
\end{tabular}
\caption{Additional experiments with a smaller GTC (Genocide Transcript Corpus) sample size to control for dataset size effects.}
\label{tab:gtc-small}
\end{table*}

%\begin{table*}[]
%    \centering
%    \begin{tabular}{lcccc}
%    \toprule
%       LM & GTC-1000 & GTC-All \\
%    \midrule
%FeedForwardModel & \wstd{0.86}{0.01}  & \wstd{0.84}{0.09} \\
%BERTPretrainedmodel & \wstd{0.93}{0.01} & \wstd{0.96}{0.01} \\
%RoBERTaPretrainedmodel & \wstd{0.95}{0.01} & \wstd{0.97}{0.01} \\
%\bottomrule
%\end{tabular}
%\caption{ROC Scores for different datasets sizes (GTC).}
%\label{tab:my_label}
%\end{table*}

\begin{figure*}[htbp]
    \centering
    \begin{subfigure}[b]{0.45\textwidth}
        \centering
        \includegraphics[width=\textwidth]{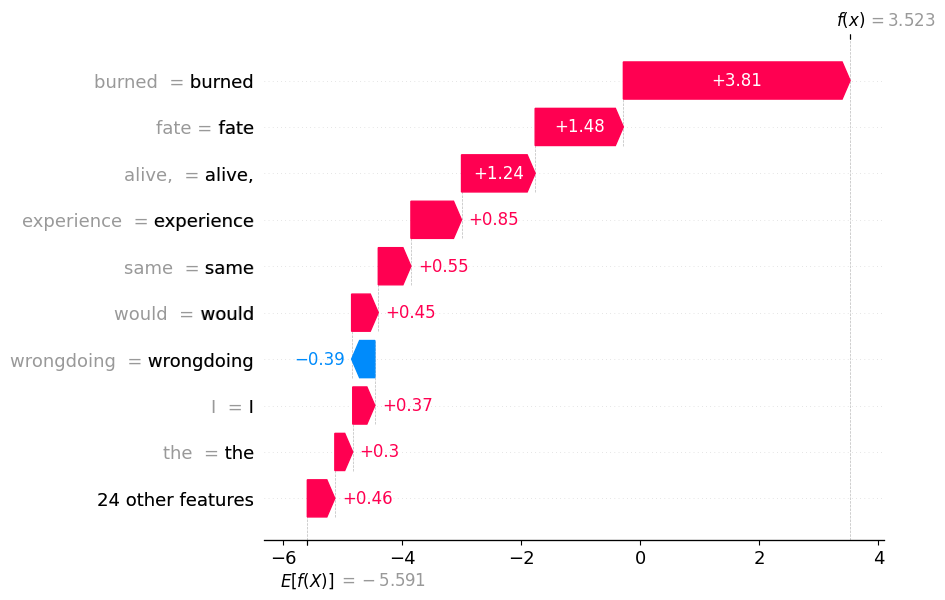}
        \caption{RoBERTa - Genocide Transcript Corpus}
        \label{fig:first_image}
    \end{subfigure}
    \hfill
    \begin{subfigure}[b]{0.45\textwidth}
        \centering
        \includegraphics[width=\textwidth]{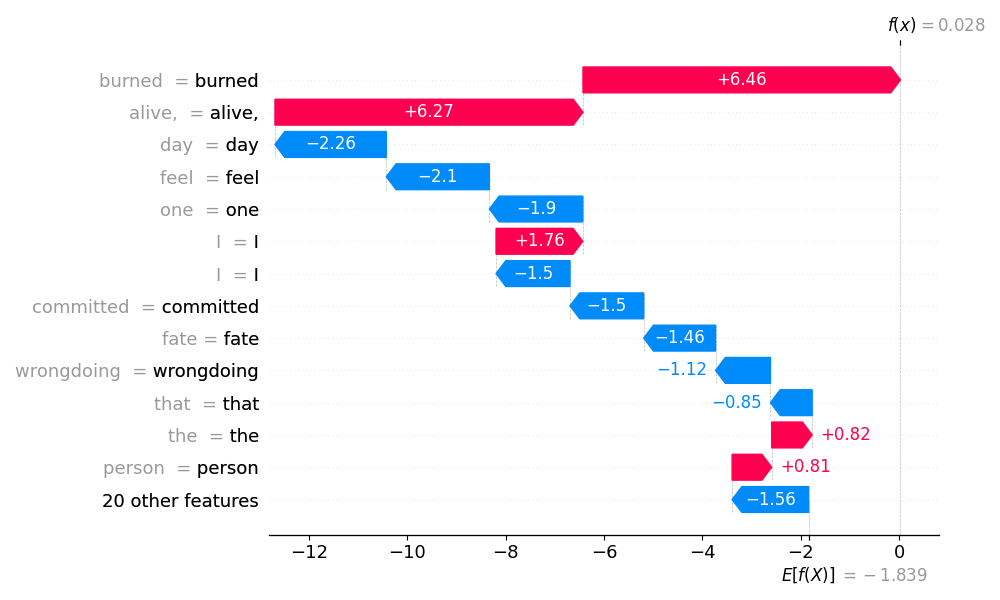}
        \caption{GPT-4 - Genocide Transcript Corpus}
        \label{fig:second_image}
    \end{subfigure}
    \vfill
    \begin{subfigure}[b]{0.45\textwidth}
        \centering
        \includegraphics[width=\textwidth]{latex/figures/shap_roberta_counsel_9.png}
        \caption{RoBERTa - Counseling Dataset (Instance 1)}
        \label{fig:third_image}
    \end{subfigure}
    \hfill
    \begin{subfigure}[b]{0.45\textwidth}
        \centering
        \includegraphics[width=\textwidth]{latex/figures/shap_gpt_counsel_9.png}
        \caption{GPT-4 - Counseling Dataset (Instance 1)}
        \label{fig:fourth_image}
    \end{subfigure}
    \vfill
    \begin{subfigure}[b]{0.45\textwidth}
        \centering
        \includegraphics[width=\textwidth]{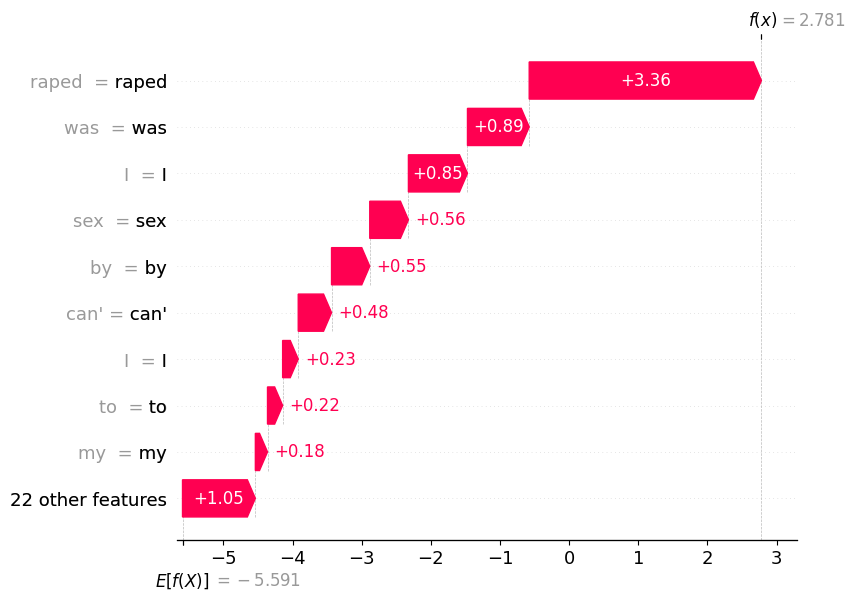}
        \caption{RoBERTa - Counseling Dataset (Instance 2)}
        \label{fig:fifth_image}
    \end{subfigure}
    \hfill
    \begin{subfigure}[b]{0.45\textwidth}
        \centering
        \includegraphics[width=\textwidth]{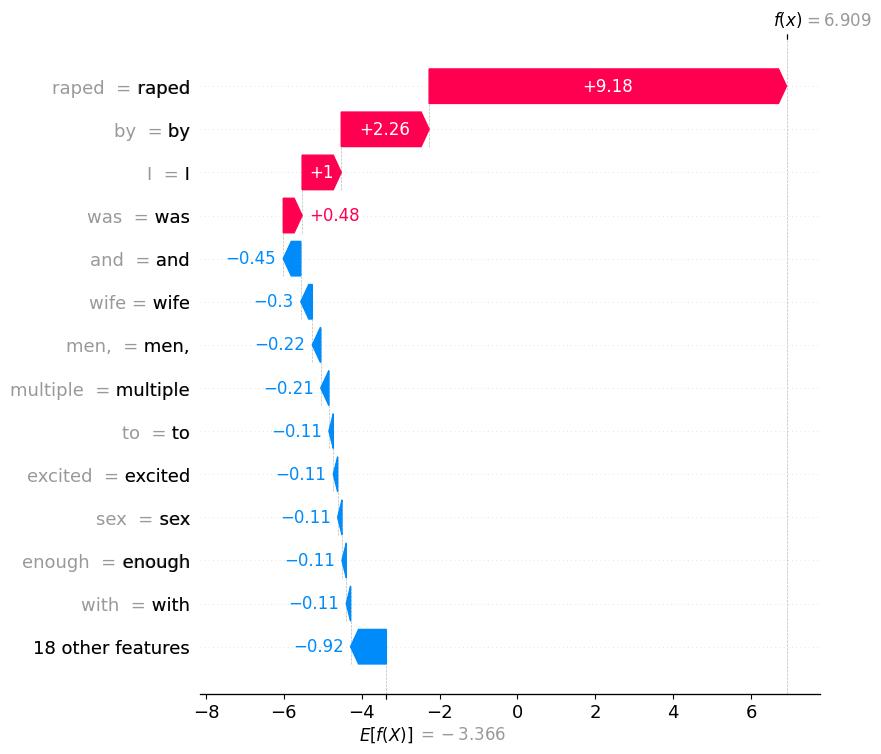}
        \caption{GPT-4 - Counseling Dataset (Instance 2)}
        \label{fig:sixth_image}
    \end{subfigure}
    \vfill
    \begin{subfigure}[b]{0.45\textwidth}
        \centering
        \includegraphics[width=\textwidth]{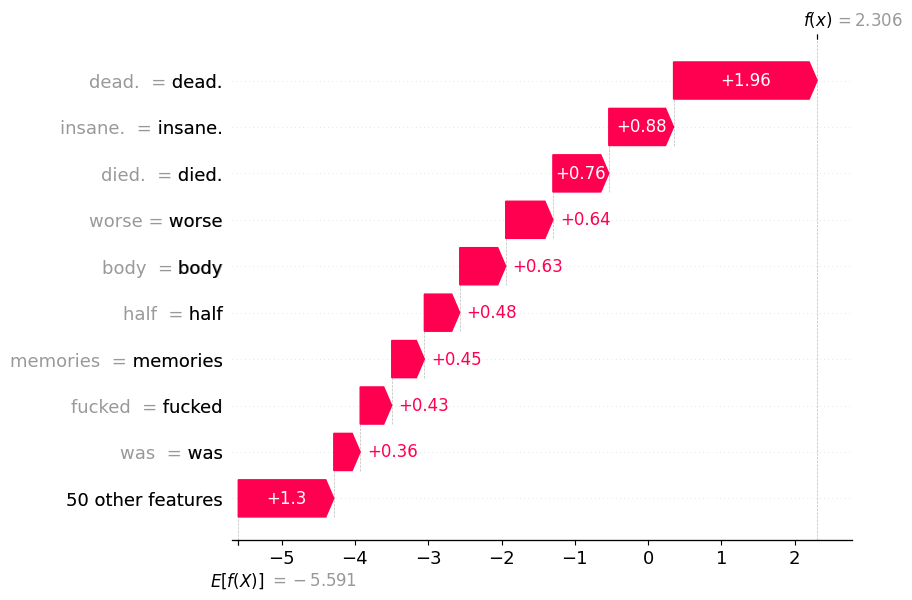}
        \caption{RoBERTa - PTSD Dataset}
        \label{fig:seventh_image}
    \end{subfigure}
    \hfill
    \begin{subfigure}[b]{0.45\textwidth}
        \centering
        \includegraphics[width=\textwidth]{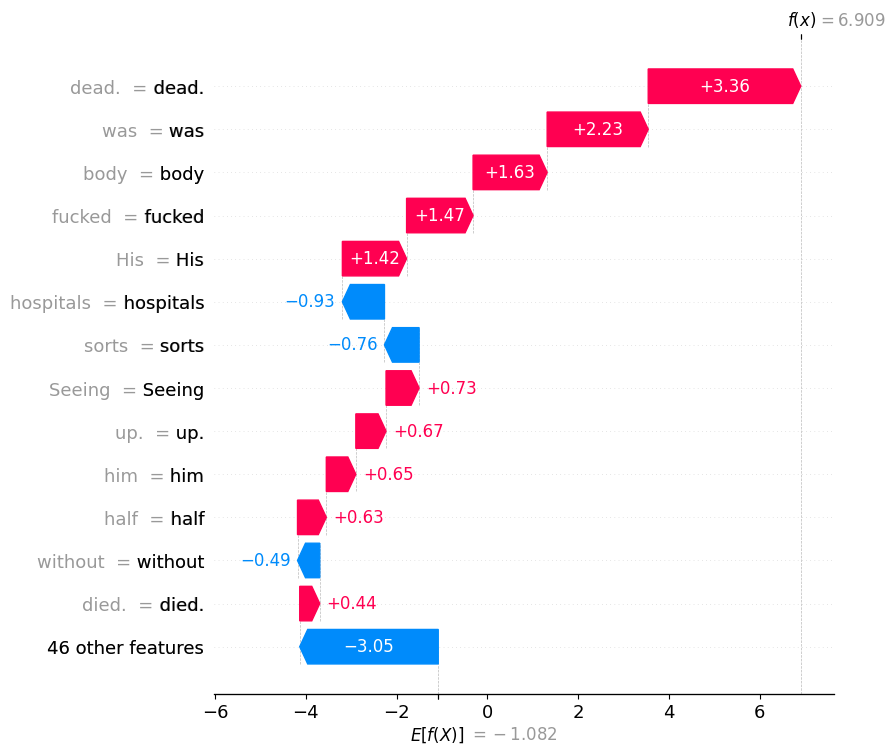}
        \caption{GPT-4 - PTSD Dataset}
        \label{fig:eighth_image}
    \end{subfigure}
    \caption{SHAP Values for various instances from different datasets. See \Cref{tab:instances} for the full text of each instance.}
    \label{fig:combined_shap_values}
\end{figure*}

\begin{table*}[htbp]
    \centering
    \begin{tabular}{|c|p{12cm}|}
        \hline
        \textbf{Dataset} & \textbf{Instance} \\
        \hline
        Genocide Transcript Corpus & I can feel that the person committed any wrongdoing would be burned alive, and I would also see that one day if I committed any wrongdoing I would experience the same fate. \\
        \hline
        Counseling Dataset (Instance 1) & My dad doesn't like the fact that I'm a boy. He yells at me daily because of it and he tells me I'm extreme and over dramatic. I get so depressed because of my dad's yelling. He keeps asking me why I can't just be happy the way I am and yells at me on a daily basis. Is this considered emotional abuse? \\
        \hline
        Counseling Dataset (Instance 2) & I was raped by multiple men, and now I can't stand the sight of myself. I wear lingerie to get my self excited enough to have sex with my wife. \\
        \hline
        PTSD Dataset & It's nearly been 4 years (trigger warning) It's almost been 4 years since he died. I can't look at hospitals without the memories coming back. Seeing him half dead. His body was all sorts of fucked up. I can't deal with this any longer. I'm going to go insane. Every day it gets worse. \\
        \hline
    \end{tabular}
    \caption{Instances from various datasets used for SHAP value analysis (see \Cref{fig:combined_shap_values}).}
    \label{tab:instances}
\end{table*}

\begin{figure*}[htbp]

\begin{minipage}[t]{\textwidth}
    \centering
    \centering
\begin{subfigure}[t]{0.24\textwidth}
\centering
    \scalebox{0.6}{
    \begin{tikzpicture}
    % Define the box
    \node[draw=brown, thick, rounded corners, text width=\boxwidth, minimum height = 9.5cm, inner sep=10pt] (box) {
        % Heading
        \centering
        \textbf{Concept 2} \\
        % Draw a horizontal line
        \tikz \draw[brown, thick] (0,0) -- (\linewidth,0);\vspace{1em}
        % Lines of text\
       \fixheightbox{
 some cows were \highlight{wounded.}\\
 or \highlight{diarrhoea}.\\
 supply themselves with food.\\
 so \highlight{itchy everywhere}.\\
 \highlight{bullet} in the head;\\
 those various \highlight{gunshots}.</s>\\
 once I was \highlight{wounded}.\\
, one \highlight{hit me} in\\
 three bursts of \highlight{gunfire}.\\
 themselves from the \highlight{bullet}s.\\
 to work I kept crying\\
 and \highlight{killed} animals for wedding\\
 the \highlight{gunshots} and we were\\
 I even saw \highlight{dead bodies}\\
I cry every night.
        }
    };
\end{tikzpicture}}
\caption{firearms, physical injuries}
\end{subfigure}
\begin{subfigure}[t]{0.24\textwidth}
\centering
    \scalebox{0.6}{
    \begin{tikzpicture}
    % Define the box
    \node[draw=brown, thick, rounded corners, text width=\boxwidth, minimum height = 9.5cm, inner sep=10pt] (box2) 
    {\centering
        % Heading
        \textbf{Concept 4} \\
        % Draw a horizontal line
        \tikz \draw[brown, thick] (0,0) -- (\linewidth,0);\vspace{1em}
        \fixheightbox{
and when he \highlight{attacked} me\\
 chief, was \highlight{very cruel}\\
 I was \highlight{punished} that way\\
  He \highlight{pressed} me against\\
 His disappearance was very \highlight{painful}\\
 Bou Meng was \highlight{tortured} for\\
 who \highlight{tortured} me was Si\\
 and then my eyes would\\
olded me for being blinded\\
 so I had him buried\\
, they stopped \highlight{beating} me\\
 who tore the child away\\
 all the \highlight{beatings} that\\
 They started \highlight{beating} me,\\
 task of killing people.
        }
    };
\end{tikzpicture}}
\caption{torture, abuse}
\end{subfigure}
\begin{subfigure}[t]{0.24\textwidth}
\centering
    \scalebox{0.6}{
    \begin{tikzpicture}
    % Define the box
    \node[draw=brown, thick, rounded corners, text width=\boxwidth, minimum height = 9.5cm, inner sep=10pt] (box2) 
    {\centering
        % Heading
        \textbf{Concept 6} \\
        % Draw a horizontal line
        \tikz \draw[brown, thick] (0,0) -- (\linewidth,0);\vspace{1em}
        \fixheightbox{
 long you \highlight{fell unconscious}.\\
 falling into the ground.\\
 another place two \highlight{bodies},\\
 them \highlight{were lying} on the\\
 weapon in my mouth.\\
 I started freezing, and\\
 up the \highlight{dead bodies} and\\
 us drowned in the river\\
 a \highlight{lot of dead} and\\
 people who \highlight{perished} there.\\
 children \highlight{die} of hunger,\\
 then we all got stuck\\
 people and throwing their \highlight{bodies}\\
 her sister burned to \highlight{death}\\
 who \highlight{died} of hunger,
        }
    };
\end{tikzpicture}}
\caption{death, motionless bodies}
\end{subfigure}
\begin{subfigure}[t]{0.24\textwidth}
\centering
    \scalebox{0.6}{
    \begin{tikzpicture}
    % Define the box
    \node[draw=brown, thick, rounded corners, text width=\boxwidth, minimum height = 9.5cm, inner sep=10pt] (box2) 
    {\centering
        % Heading
        \textbf{Concept 7} \\
        % Draw a horizontal line
        \tikz \draw[brown, thick] (0,0) -- (\linewidth,0);\vspace{1em}
        \fixheightbox{
December he \highlight{called} me and\\
 my husband \highlight{called} cadres\\
 sometimes he  \highlight{called} me to\\
 his phone  \highlight{call} because he\\
 or  \highlight{called} me and then\\
 was  \highlight{called} Lucia, Ruk\\
 know what happened to Ph\\
 man  \highlight{called} Rukara went\\
 since I was busy looking\\
 school was  \highlight{called} Hasan Ve\\
 really is still in my\\
 one \highlight{called} me up and\\
 my wife  \highlight{asked} them what\\
  \highlight{tell} you exactly when this\\
 It arrived in Kosovo Pol
        }
    };
\end{tikzpicture}}
\caption{communication, asking questions (non-traumatic)}
\end{subfigure}
    \caption*{(a) Concepts Found in the \textbf{GTC Dataset} (Examples, RoBERTa Model): (a)-(c) Trauma-Related, (d) Non-Trauma Related}
\end{minipage}

\begin{minipage}[t]{\textwidth}
    \centering
    \centering
\begin{subfigure}[t]{0.24\textwidth}
\centering
    \scalebox{0.6}{
    \begin{tikzpicture}
    % Define the box
    \node[draw=brown, thick, rounded corners, text width=\boxwidth, minimum height = 9.5cm, inner sep=10pt] (box) {
        % Heading
        \centering
        \captionsetup{justification=raggedright, singlelinecheck=false}
        \textbf{Concept 0} \\
        % Draw a horizontal line
        \tikz \draw[brown, thick] (0,0) -- (\linewidth,0);\vspace{1em}
        % Lines of text\
       \fixheightbox{
 in the \highlight{dream} with the\\
In this \highlight{dream},\\
 awake in the \highlight{dream},\\
I survived \highlight{veteran} suicide.\\
 Or in a \highlight{dream} rel\\
 toxic relationships with men.\\
,500 \highlight{military} sexual assaults\\
 \highlight{shot and killed himself} in\\
 ago I had a \highlight{dream}\\
 only dated for a few\\
 after having a \highlight{dream} that\\
harmed while in a\\
 was a \highlight{dream}. This\\
 \highlight{shot herself} in the head\\
, on a medical discharge
        }
    };
\end{tikzpicture}}
\caption{dreams, military, suicide}
\end{subfigure}
\begin{subfigure}[t]{0.24\textwidth}
\centering
    \scalebox{0.6}{
    \begin{tikzpicture}
    % Define the box
    \node[draw=brown, thick, rounded corners, text width=\boxwidth, minimum height = 9.5cm, inner sep=10pt] (box2) 
    {\centering
        % Heading
        \textbf{Concept 5} \\
        % Draw a horizontal line
        \tikz \draw[brown, thick] (0,0) -- (\linewidth,0);\vspace{1em}
        \fixheightbox{
. Our \highlight{car} clipped the\\
 his \highlight{car} multiple times.\\
 my \highlight{skull}. It wasn\\
 \highlight{raped} at least three times\\
 hit another \highlight{car}, causing\\
 22 year old man (\\
 were later found and her\\
 a lot of \highlight{death},\\
\highlight{SA} and \highlight{death}\\
 of \highlight{sexual abuse}, two\\
 \highlight{murder-suicides} each\\
 tried to apprehend the two\\
/SuicideWatch and\\
 after service. Fires are\\
 experiences with SSRIs
        }
    };
\end{tikzpicture}}
\caption{accidents, death, sexual abuse}
\end{subfigure}
\begin{subfigure}[t]{0.24\textwidth}
\centering
    \scalebox{0.6}{
    \begin{tikzpicture}
    % Define the box
    \node[draw=brown, thick, rounded corners, text width=\boxwidth, minimum height = 9.5cm, inner sep=10pt] (box2) 
    {\centering
        % Heading
        \textbf{Concept 6} \\
        % Draw a horizontal line
        \tikz \draw[brown, thick] (0,0) -- (\linewidth,0);\vspace{1em}
        \fixheightbox{
 my \highlight{parents} were kids,\\
 my \highlight{childhood} over to my\\
 my \highlight{mom divorced} my violent\\
 I was a \highlight{child},\\
 had an abusive ex and\\
 my \highlight{mother} died 2 years\\
 memories of my \highlight{childhood},\\
 abusive and cheating gay ex\\
 first serious boyfriend repeatedly d\\
 realize TW: \highlight{childhood} sexual\\
 my trauma from \highlight{childhood},\\
 my now ex of 4\\
 My \highlight{parents} were divorced when\\
 my \highlight{parents} raised me and\\
 throughout my \highlight{childhood} by my
        }
    };
\end{tikzpicture}}
\caption{childhood abuse, family trauma}
\end{subfigure}
\begin{subfigure}[t]{0.24\textwidth}
\centering
    \scalebox{0.6}{
    \begin{tikzpicture}
    % Define the box
    \node[draw=brown, thick, rounded corners, text width=\boxwidth, minimum height = 9.5cm, inner sep=10pt] (box2) 
    {\centering
        % Heading
        \textbf{Concept 9} \\
        % Draw a horizontal line
        \tikz \draw[brown, thick] (0,0) -- (\linewidth,0);\vspace{1em}
        \fixheightbox{
 extremely frequent \highlight{flashbacks} the \\
 me bad. The \highlight{flashbacks}\\
 When I was \highlight{molested}\\
 have vivid \highlight{flashbacks}. All\\
\highlight{aphobia} in the last\\
 about \highlight{flashbacks}.
I\\
 \highlight{phobia} related to the\\
 after i was \highlight{sexually assaulted}\\
 young child, was sexually\\
 having nightmares and \highlight{flashbacks}\\
 \highlight{repressed memories} are a\\
 because I have \highlight{flashbacks} several\\
 always thought the \highlight{memories}\\
 of scolding via email\\
 like I was \highlight{abused}.
        }
    };
\end{tikzpicture}}
\caption{flashbacks, memories, sexual abuse, phobia}
\end{subfigure}
    \caption*{(b) Concepts Found in the \textbf{PTSD Dataset} (Examples, RoBERTa Model): (a)-(d) Trauma-Related}
\end{minipage}

\begin{minipage}[t]{\textwidth}
    \centering
    \centering
\begin{subfigure}[t]{0.24\textwidth}
\centering
    \scalebox{0.6}{
    \begin{tikzpicture}
    % Define the box
    \node[draw=brown, thick, rounded corners, text width=\boxwidth, minimum height = 9.5cm, inner sep=10pt] (box) {
        % Heading
        \centering
        \textbf{Concept 4} \\
        % Draw a horizontal line
        \tikz \draw[brown, thick] (0,0) -- (\linewidth,0);\vspace{1em}
        % Lines of text\
       \fixheightbox{
 My boyfriend lost his \highlight{dad}\\
 my \highlight{mom}. My \highlight{dad}\\
 shot and killed my rapist\\
 His \highlight{mom}, my \highlight{grandma}\\
My \highlight{dad} cheated on my\\
 I lost my \highlight{mother} recently\\
 his \highlight{wife}. My \highlight{uncle}\\
, my \highlight{mother} and \highlight{father}\\
I was kidnapped at fourteen\\
. My \highlight{father} cheated on\\
My ex-wife married\\
 \highlight{niece} whom my \highlight{sister} abandoned\\
My \highlight{daughter} was overly tired\\
 my \highlight{mom}. Years later\\
 misses \highlight{mom} and \highlight{dad} in
        }
    };
\end{tikzpicture}}
\caption{family members: cheating, loss}
\end{subfigure}
\begin{subfigure}[t]{0.24\textwidth}
\centering
    \scalebox{0.6}{
    \begin{tikzpicture}
    % Define the box
    \node[draw=brown, thick, rounded corners, text width=\boxwidth, minimum height = 9.5cm, inner sep=10pt] (box2) 
    {\centering
        % Heading
        \textbf{Concept 5} \\
        % Draw a horizontal line
        \tikz \draw[brown, thick] (0,0) -- (\linewidth,0);\vspace{1em}
        \fixheightbox{
My boyfriend lost his dad\\
 my mom. My dad\\
 shot and killed my rapist\\
 his wife. My uncle\\
 His mom, my grandma\\
My dad cheated on my\\
 I lost my mother recently\\
, my mother and father\\
. My father cheated on\\
My ex-wife married\\
I was kidnapped at fourteen\\
\highlight{Post Traumatic Stress Disorder}\\
\highlight{Post Traumatic Stress Disorder}\\
niece whom my sister abandoned\\
\highlight{posttraumatic stress disorder}
        }
    };
\end{tikzpicture}}
\caption{PTSD (overlap Concept 4)}
\end{subfigure}
\begin{subfigure}[t]{0.24\textwidth}
\centering
    \scalebox{0.6}{
    \begin{tikzpicture}
    % Define the box
    \node[draw=brown, thick, rounded corners, text width=\boxwidth, minimum height = 9.5cm, inner sep=10pt] (box2) 
    {\centering
        % Heading
        \textbf{Concept 6} \\
        % Draw a horizontal line
        \tikz \draw[brown, thick] (0,0) -- (\linewidth,0);\vspace{1em}
        \fixheightbox{
 My boyfriend lost his dad\\
 my mom. My dad\\
 abusive father and his wife\\
 his wife. My uncle\\
 His mom, my grandma\\
My dad cheated on my\\
 I lost my mother recently\\
, my mother and father\\
I was kidnapped at fourteen\\
 My sister \highlight{never defended me}\\
, my \highlight{doctor gave me}\\
. He \highlight{bought me} a\\
 all of my family left\\
 \highlight{broken apart} after she got\\
 \highlight{rehabilitation} program and got kicked
        }
    };
\end{tikzpicture}}
\caption{clinical context, dependencies (overlap Concept 4)}
\end{subfigure}
\begin{subfigure}[t]{0.24\textwidth}
\centering
    \scalebox{0.6}{
    \begin{tikzpicture}
    % Define the box
    \node[draw=brown, thick, rounded corners, text width=\boxwidth, minimum height = 9.5cm, inner sep=10pt] (box2) 
    {\centering
        % Heading
        \textbf{Concept 9} \\
        % Draw a horizontal line
        \tikz \draw[brown, thick] (0,0) -- (\linewidth,0);\vspace{1em}
        \fixheightbox{
I was violently \highlight{raped} by\\
 got \highlight{pregnant} by my boyfriend\\
 my \highlight{baby} mother. She\\
 my children's father left\\
I saw my mother cheating\\
I was \highlight{raped} by multiple\\
My girlfriend was \highlight{abused} as\\
 I got \highlight{raped} by my\\
 I just lost my mom\\
 teenager. My entire family\\
I was \highlight{raped} repeatedly when\\
My grandma and brother both\\
 parents injured my brother,\\
, my husband mentally abused\\
My mother has Alzheimer's
        }
    };
\end{tikzpicture}}
\caption{rape, abuse, pregnancy}
\end{subfigure}
    \caption*{(c) Concepts Found in the \textbf{Counseling Dataset} (Examples, RoBERTa Model): (a)-(d) Trauma-Related}
\end{minipage}

\caption{Examples of Concepts Discovered on Various Datasets for the RoBERTa Model: (a) GTC dataset, (b) PTSD dataset, (c) Counseling dataset.}
\label{fig:conceptstraumagtc_combined}
\end{figure*}

\end{document}